\documentclass[sigconf, timestamp=false]{acmart}

\AtBeginDocument{%
  }

\acmISBN{978-1-4503-XXXX-X/2018/06}

\usepackage{amsmath,amsfonts}
\usepackage{algorithmic}
\usepackage{graphicx}
\usepackage{textcomp}
\usepackage{xcolor}
\usepackage{booktabs}
\usepackage{comment}
\usepackage{float}
\usepackage[final]{pdfpages}
\usepackage{hyperref}
\usepackage{url}

\usepackage{breakurl}
\usepackage{tablefootnote}
\usepackage{subcaption}
\usepackage{multirow}
\usepackage{caption}
\usepackage{subcaption}
\usepackage{amsmath}
\usepackage{flushend}
\usepackage{balance}
\usepackage{ulem}
\usepackage{svg}
\usepackage{comment}
\usepackage[bottom]{footmisc}
\usepackage{nameref}
\usepackage[switch]{lineno}

\usepackage{nicematrix}
\usepackage[utf8]{inputenc}
\usepackage{nicematrix}
\usepackage[T1]{fontenc}
\usepackage{mathptmx}
\usepackage{etoolbox}
\usepackage{amsmath,amsfonts}
\usepackage{algorithmic}
\usepackage{textcomp}
\usepackage{booktabs}
\usepackage{comment}
\usepackage{float}
\usepackage{enumitem}
\usepackage[final]{pdfpages}
\usepackage{tablefootnote}
\usepackage{caption}
\usepackage{subcaption}
\usepackage{amsmath}
\usepackage{flushend}
\usepackage{balance}
\usepackage{ulem}
\usepackage[bottom]{footmisc}
\usepackage{tabularx}
\usepackage{colortbl}
\usepackage{xargs}
\usepackage[justification=centering]{caption}

\usepackage{multirow}

\renewcommand\footnotetextcopyrightpermission[1]{}
\begin{document}

\title{MoE-Inference-Bench: Performance Evaluation of Mixture of Expert Large Language and Vision Models}

\author{
Krishna Teja Chitty-Venkata\textsuperscript{1}, 
Sylvia Howland\textsuperscript{2},
Golara Azar\textsuperscript{2},
Daria Soboleva\textsuperscript{2},
Natalia Vassilieva\textsuperscript{2}, 
Siddhisanket Raskar\textsuperscript{3},
Murali Emani\textsuperscript{1},
Venkatram Vishwanath\textsuperscript{1} \\
\textsuperscript{1}Argonne National Laboratory,
\textsuperscript{2}Cerebras,
\textsuperscript{3}Pacific Northwest National Laboratory\\
{
\{schittyvenkata, memani, venkat\}@anl.gov,
s.raskar@pnnl.gov, \\
\{sylvia.howland, golara.azar, daria.soboleva, natalia\}@cerebras.net
}
}




\begin{abstract}

Mixture of Experts (MoE) models have enabled the scaling of Large Language Models (LLMs) and Vision Language Models (VLMs) by achieving massive parameter counts while maintaining computational efficiency. However, MoEs introduce several inference-time challenges, including load imbalance across experts and the additional routing computational overhead. To address these challenges and fully harness the benefits of MoE, a systematic evaluation of hardware acceleration techniques is essential. We present MoE-Inference-Bench, a comprehensive study to evaluate MoE performance across diverse scenarios. We analyze the impact of batch size, sequence length, and critical MoE hyperparameters such as FFN dimensions and number of experts on throughput. We evaluate several optimization techniques on Nvidia H100 GPUs, including pruning, Fused MoE operations, speculative decoding, quantization, and various parallelization strategies. Our evaluation includes MoEs from the Mixtral, DeepSeek, OLMoE and Qwen families. The results reveal performance differences across configurations and provide insights for the efficient deployment of MoEs.
\end{abstract}

\newcommand {\ME}[1]{{\textcolor{blue}{(Murali: #1)}}}


\setcopyright{none}

\settopmatter{printfolios=false,printccs=false,printacmref=false}
\renewcommand\footnotetextcopyrightpermission[1]{}
\fancyhead{}

\maketitle

\section{Introduction}
\label{sec:intro}


\textit{Mixture of Experts} (MoE) models have emerged as a powerful paradigm for scaling neural networks, particularly in the domain of Large Language Models (LLMs). This approach offers a way to increase model capacity without a proportional rise in computational cost.
MoE differs from dense models by using multiple specialized sub-networks, where each input activates only a subset of experts (as determined by a gating network). In contrast, dense models activate all parameters for every input, making MoE architectures significantly more parameter-efficient. Architectures such as the Switch Transformer \cite{fedus2022switch} and GShard \cite{lepikhin2020gshard}, along with more recent open-source MoE models like Mixtral \cite{jiang2024mixtral}, Llama4 \cite{meta2025llama}, DeepSeekMoE \cite{dai2024deepseekmoe}, and Kimi \cite{team2025kimi}, exemplify the rapid advancements in MoE-based systems
These models leverage sparse weight activation, enabling large networks to maintain inference efficiency. MoE models are now widely used in applications such as text generation, retrieval-augmented generation, and multimodal reasoning. However, despite their computational advantages, MoE models also pose unique challenges in inference, training stability, memory usage, and hardware utilization due to load imbalance and dynamic routing.

\begin{figure}
    \centering
    \includegraphics[width=0.9\linewidth]{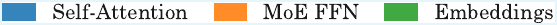}
    \begin{subfigure}{0.3\linewidth}
        \centering
        \includegraphics[width=\linewidth]{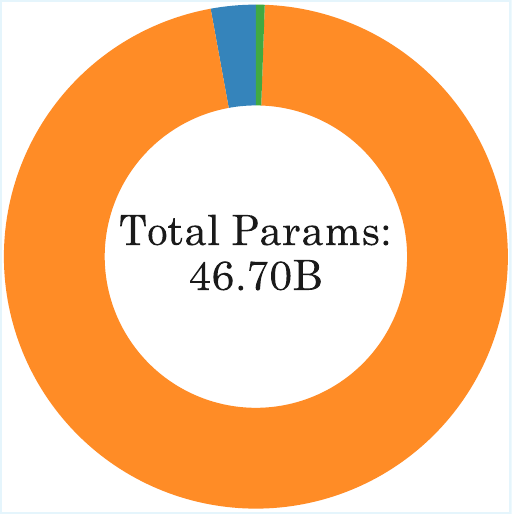}
        \caption{Mixtral-8x7B}
    \end{subfigure}
    \begin{subfigure}{0.3\linewidth}
        \centering
        \includegraphics[width=\linewidth]{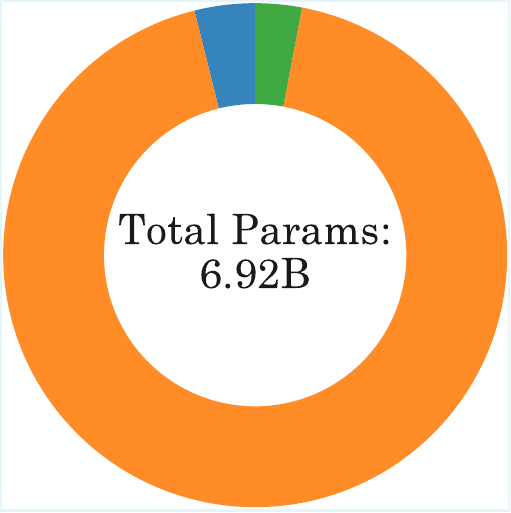}
        \caption{OLMoE-1B-7B}
    \end{subfigure}
    \begin{subfigure}{0.3\linewidth}
        \centering
        \includegraphics[width=\linewidth]{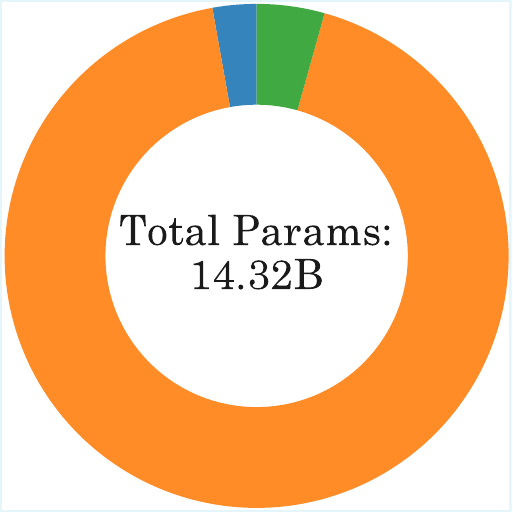}
        \caption{Qwen1.5-MoE}
    \end{subfigure}
   \begin{subfigure}{0.3\linewidth}
        \centering
        \includegraphics[width=\linewidth]{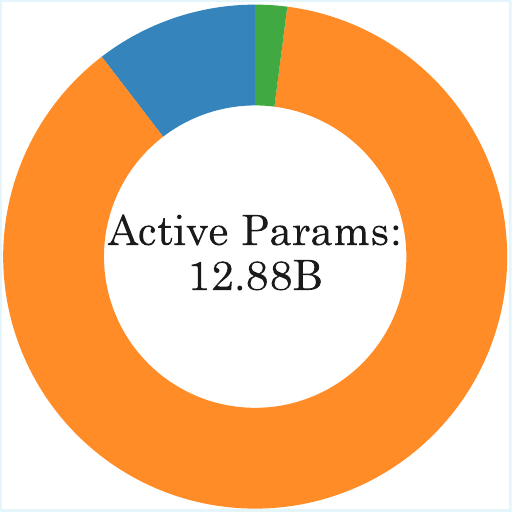}
        \caption{Mixtral-8x7B}
    \end{subfigure}
    \begin{subfigure}{0.3\linewidth}
        \centering
        \includegraphics[width=\linewidth]{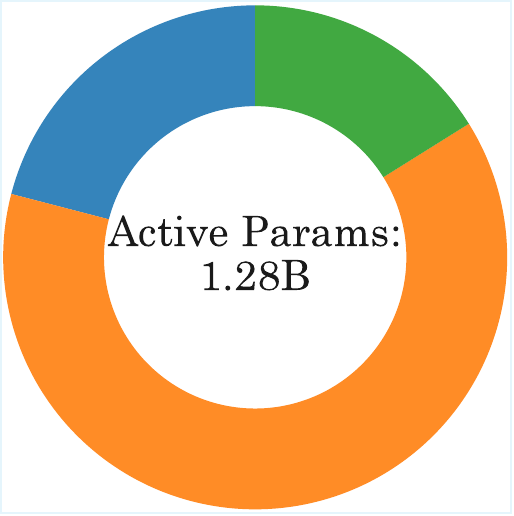}
         \caption{OLMoE-1B-7B}
    \end{subfigure}
    \begin{subfigure}{0.3\linewidth}
        \centering
        \includegraphics[width=\linewidth]{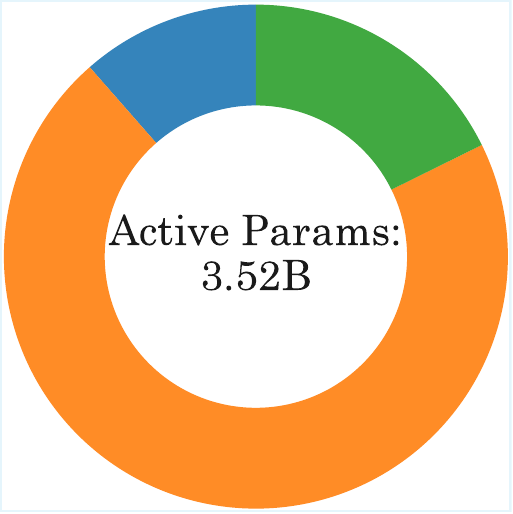}
        \caption{Qwen1.5-MoE}
    \end{subfigure}
    \vspace{-3mm}
    \caption{Layer-wise Total and Active Parameter Breakdown for Mixtral-8x7B, OLMoE-1B-7B, and Qwen1.5-MoE}
    \label{fig:parameters}
    \vspace{-5mm}
\end{figure}

\textit{MoE Inference} \cite{liu2024survey} plays a central role in modern AI applications, as it involves executing the forward pass of a sparsely activated model where only the top-$k$ experts per token are evaluated. Efficient inference is crucial for maximizing the benefits of sparsity in real-world deployments. As MoE models continue to grow in scale and complexity, optimizing inference is critical to achieve low latency and energy-efficien execution on modern accelerators. This includes mitigating expert load imbalance, reducing communication overhead in distributed settings, and designing scheduling strategies that fully exploit sparsity for throughput gains.

The MoE ecosystem has witnessed a convergence of three key trends: the rise of open-source MoE models, advancements in AI accelerators, and the development of inference frameworks like vLLM \cite{kwon2023efficient} and FasterMoE \cite{he2022fastermoe} optimized for sparse execution. This synergy highlights the importance of robust benchmarking to evaluate MoE performance across diverse hardware setups. Benchmarking exposes critical trade-offs between throughput, latency, and memory footprint, enabling informed decisions about model deployment and architecture optimization.

The evolution of AI hardware such as GPUs and specialized AI accelerators has been instrumental for the ever-rising computational demands of MoE models. These accelerators offer high parallelism and memory bandwidth, essential for models with billions of parameters and dynamic computation graphs. However, MoE architectures also introduce new hardware challenges, such as expert placement, routing overhead, and under-utilization due to sparse activations. Addressing these hardware inefficiencies requires co-designing inference systems that are both MoE-aware and hardware-efficient. Figure \ref{fig:parameters} shows that MoE layers dominate both total and active parameters across different models, emphasizing their critical role in computational cost and memory footprint. Since MoE layer weights account for a substantial portion of the model, understanding the MoE performance is essential for optimized deployment.

In this paper, we introduce MoE-Inference-Bench, a comprehensive benchmarking suite designed to systematically evaluate MoE models across a wide range of optimization techniques. Our benchmark analyzes throughput, latency, and hardware utilization for state-of-the-art MoE models, shedding light on the practical implications of sparse inference and routing dynamics. Our comprehensive study provides several insights for researchers aiming to deploy MoE models efficiently, and contributes to the broader goal of scalable and cost-effective AI deployment in the era of massive model sparsity.

The main contributions of our paper are as follows: 

\begin{enumerate} 
    
    
    
\item \textbf{Comprehensive MoE Benchmarking Suite:} We propose MoE-Inference-Bench to evaluate MoE performance under diverse inference scenarios. Our suite spans models from 2B to 70B parameters, covering multiple architectures (Mixtral, DeepSeek, Qwen, Phi, OLMoE). Our study examines multiple factors that significantly influence the inference performance of MoEs, providing insights for future designs. 

\item \textbf{Fine-Grained Hyperparameter Scaling Analysis:} We perform an extensive exploration of key MoE layer hyperparameters, which include FFN dimension, total expert count, and active expert ratio to quantify their individual and joint impact on throughput and out-of-memory boundaries on Nvidia H100 GPUs. Our results identify optimal MoE operating constraints and reveal clear trade-offs between model size, expert sparsity and hardware efficiency.

\item \textbf{Inference Optimizations:} We systematically assess multiple inference-time acceleration techniques such as quantization, intra and inter expert pruning, speculative decoding and Fused MoE, highlighting their effectiveness across batch sizes and sequence lengths. We also benchmark MoE inference across Nvidia H100 GPUs, analyzing the effects of tensor, pipeline, and expert parallelism strategies. 

\end{enumerate}

\section{Background and Related Work}
\label{sec:background}


\paragraph{Large Language and Vision Models} Modern LLMs are predominantly built upon the transformer architecture \cite{vaswani2017attention}, which comprises stacks of decoder layers. These layers incorporate core components such as token embeddings, positional encodings, multi-head self-attention, and feed-forward networks. VLMs combine vision and language capabilities to simultaneously process both visual data and textual information, enabling them to perform multimodal tasks such as image captioning and visual question answering. 

\begin{figure}
    \centering
    \begin{subfigure}{\linewidth}
        \centering
        \includegraphics[width=0.8\linewidth]{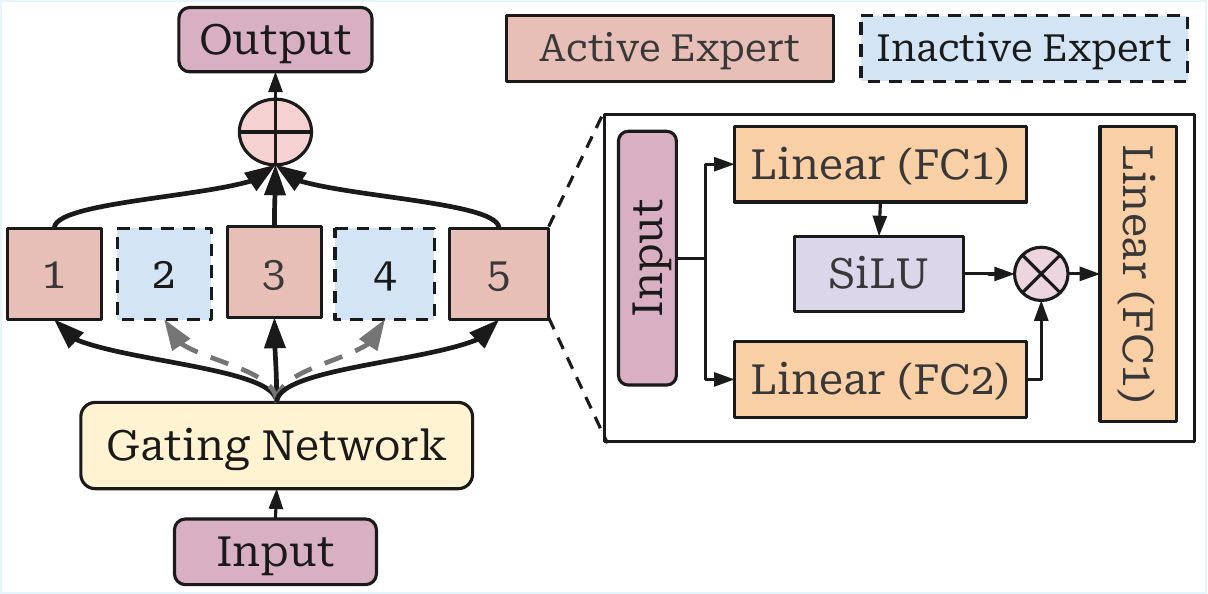}
    \end{subfigure}
    \caption{Mixture of Expert (MoE) Design}
    \label{fig:moe_design_background}
    \vspace{-5mm}
\end{figure}

\paragraph{{Mixture-of-Experts LLMs} }
Dense architectures represent the conventional LLM, where a single, monolithic neural network activates all parameters for every token. This design facilitates comprehensive information processing but incurs substantial computational and memory costs \cite{touvron2023llama}. Mixture-of-Experts (MoE) models \cite{shazeer2017outrageously, cai2024survey} incorporates multiple specialized subnetworks within selected layers, typically the FFN blocks, as shown in Figure \ref{fig:moe_design_background}. A learnable routing mechanism activates only a subset of experts per token, improving parameter efficiency and potentially accelerating inference without proportionally increasing compute. Notable Examples include Mixtral-8x7B \cite{mixtral_8_7b}, where expert specialization enables scaling to larger total parameter counts while mitigating the runtime overhead of dense activation. However, MoE architectures introduce additional complexity in training stability and load balancing.

\paragraph{Benchmarking LLM Performance}
LLM Benchmarking under different optimizations is essential for assessing the computational trade-offs of diverse architectures. Previous studies have evaluated LLMs on leadership-class supercomputers \cite{emani2023gpt2, llmonfrontier}, LLM-specific inference \cite{chitty2024llm} and deep learning benchmark suites \cite{emani2022cnn, YIN2021100005}, offering insights into scalability, efficiency, and hardware utilization patterns. To the best of our knowledge, this work is the first to present systematic, inference-focused benchmarking of state-of-the-art MoE models across a broad spectrum of optimizations, providing insights into architectural and system-level performance trade-offs.

\section{Experimental Setup} \label{sec:setup}


\subsection{\textbf{LLM Architectures}} 


We evaluate MoEs across varying sizes and architectures, enabling a comprehensive inference performance comparison. The models include Mixtral-8×7B \cite{jiang2024mixtral}, Qwen-1.5-MoE \cite{bai2023qwen}, Qwen3-30B-A3B \cite{yang2025qwen3}, DeepSeek-V2-Lite \cite{liu2024deepseek}, Phi-3.5-MoE \cite{abdin2024phi}, OLMoE-1B-7B \cite{muennighoff2024olmoe}, DeepSeek-VL2-Tiny, DeepSeek-VL2-Small, DeepSeek-VL2 \cite{wu2024deepseek}. This set encompasses both LLM and VLM MoEs, covering parameter scales from lightweight 7B parameter models to large-scale 30B+ parameter networks. 
Table \ref{tab:model_architectures} summarizes the architecture specifications of different MoE models in our evaluation.

\begin{table*}
\centering
\small
\caption{Comparison of Mixture of Expert Model Architectures}
\resizebox{\textwidth}{!}{
\begin{tabular}{|c|c|c|c|c|c|c|c|c|c|}
\hline
Model              & Model Type                                                               & Modality     & \#Layers & \begin{tabular}[c]{@{}c@{}}\#Hidden\ Size\end{tabular} & \begin{tabular}[c]{@{}c@{}}\#FFN\ Dimension\end{tabular} & \#Experts & \begin{tabular}[c]{@{}c@{}}\#Active\ Experts\end{tabular} & \begin{tabular}[c]{@{}c@{}}Model\ Size\end{tabular} & \begin{tabular}[c]{@{}c@{}}Active\ Parameters\end{tabular} \\
\hline
Mixtral 8x7B & Transformer & Text & 32 & 4096 & 14336 & 8 & 2 & 47B & 12.9B \\
\hline
Qwen 1.5 MoE & Transformer & Text & 24 & 2048 & 5632 & 60 & 4 & 14.3B & 2.7B \\
\hline
Qwen3-30B-A3B & Transformer & Text & 48 & 5120 & 13824 & 128 & 8 & 30.5B & 3.3B \\
\hline
DeepSeek V2 Lite & Transformer & Text & 27 & 2048 & 1408 & 64 & 6 & 15.7B & 2.4B \\
\hline 
Phi 3.5 MoE & Transformer & Text & 32 & 4096 & 6400 & 16 & 2 & 41.9B & 6.6B \\
\hline
OLMoE-1B-7B & Transformer & Text & 16 & 2048 & 8192 & 64 & 8 & 7.2B & 1.3B \\
\hline
DeepSeek VL2 Tiny & Transformer & Text + Image & 16 & 1536 & 8960 & 8 & 2 & 3B & 1.0B \\
\hline
DeepSeek VL2 Small & Transformer & Text + Image & 24 & 2048 & 11008 & 8 & 2 & 16B & 2.8B \\
\hline
DeepSeek VL2 & Transformer & Text + Image & 32 & 4096 & 14336 & 8 & 2 & 27B & 4.5B \\
\hline
\end{tabular}
}
\label{tab:model_architectures}
\end{table*}

\subsection{\textbf{LLM Token Generation Parameters}} 
The input length refers to the number of tokens present in a single input prompt. The output size denotes the number of tokens generated by the model sequentially until a stopping criterion or predefined token limit is reached. The batch size corresponds to the number of input \& output pairs processed concurrently. In our evaluation, we consider input and output lengths of 128, 256, 512, 1024, and 2048 tokens, and batch sizes of 1, 16, 32, and 64.

\subsection{\textbf{AI Hardware Platforms}} 
We deploy MoEs on Nvidia H100 SXM5 80GB GPU \cite{choquette2023nvidia} using the vLLM \cite{kwon2023efficient} framework.
The NVIDIA H100 GPU, built on TSMC’s 4N process with 80B transistors, optimized for trillion-parameter LLMs. It features 80 GB HBM3 memory, 50 MB L2 cache, fourth-generation Tensor Cores and NVLink. vLLM \cite{kwon2023efficient} is an open-source inference framework known for its efficient memory management and support across a wide range of AI accelerators. In a limited study of Llama-4 Scout performance, we include a publicly-available Cerebras cloud inference CS-3 model replica \cite{cerebras} in our evaluation. 



\subsection{\textbf{Performance Metrics}} 

We employ the following performance metrics in our evaluation:
\textbf{(a) Time to First Token (TTFT)} measures the time between receiving an input prompt and generating the first output token. It reflects the responsiveness of an LLM from the user's perspective. TTFT is obtained by limiting the maximum output length to a single token and recording the generation time.

\textbf{(b) Inter-Token Latency (ITL)} is the average time interval between producing consecutive output tokens. It captures the model's per-token generation speed. ITL is computed as:
\begin{equation} \label{eq:ITL}
\text{ITL} = \frac{\text{End-to-End Latency} - \text{TTFT}}{\text{Batch Size} \times (\text{Output Tokens} - 1)}
\end{equation}
\textbf{(c) Throughput} indicates the processing efficiency of the hardware, representing the total number of tokens (input and output) processed per second. We first measure the end-to-end latency (time from prompt submission to the generation of the final output token) and convert it to throughput as follows:
\begin{equation} \label{eq:throughput}
\text{Throughput} = \frac{\text{Batch Size} \times \textbf{(}\text{Input Tokens} \textbf{+} \text{Output Tokens}\textbf{)}}{\text{End-to-End Inference Latency}}
\end{equation}
(d) \textbf{Samples per second:}  The metric for VLMs is the number of input (image + text) samples processed per second.





\section{MoE Inference Analysis} \label{sec:infr_prelim_study}

In this section, we first examine the breakdown of MoE prefill and decode phases, followed by examining the role of input/output lengths and batch sizes on the inference throughput.

\subsection{Prefill and Decode Breakdown}

\begin{figure}
    \centering
    \includegraphics[width=\linewidth]{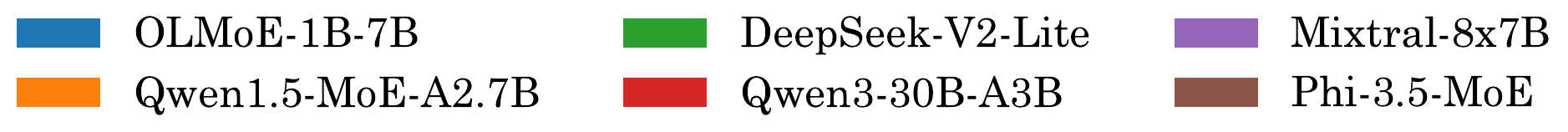}
    \begin{subfigure}{0.3\linewidth}
        \centering
        \includegraphics[width=\linewidth]{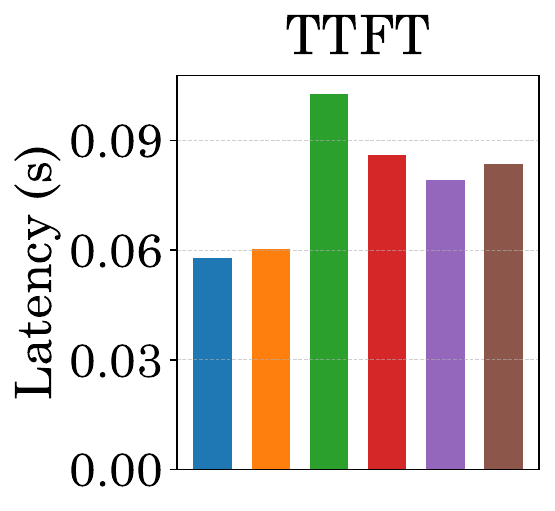}
    \end{subfigure}
    \hfill
    \begin{subfigure}{0.3\linewidth}
        \centering
        \includegraphics[width=\linewidth]{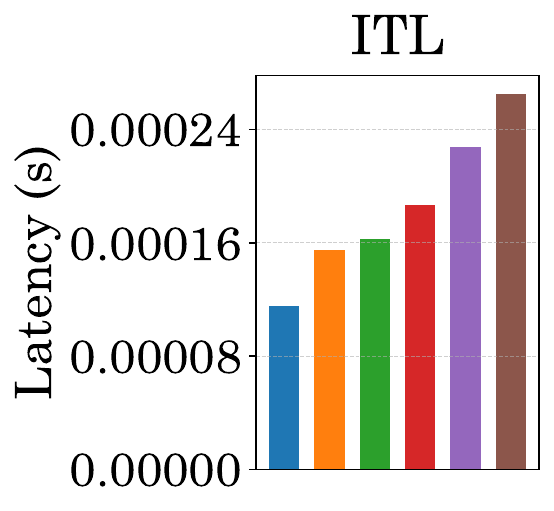}
    \end{subfigure}
    \hfill
    \begin{subfigure}{0.3\linewidth}
        \centering
        \includegraphics[width=\linewidth]{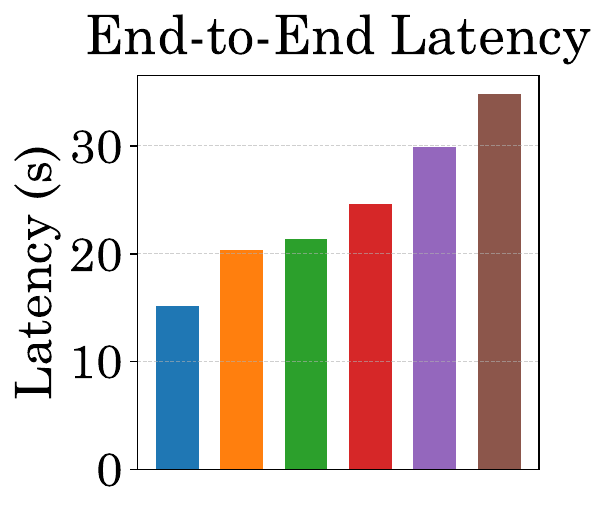}
    \end{subfigure}
    \vspace{-4mm} 
    \caption{TTFT, ITL and End-to-End Latency of LLMs for Batch Size of 64 and Input \& Output Length of 2048}
    \label{fig:LLM_prefill_decode}
    \vspace{-5mm}
\end{figure}

\begin{figure}
    \centering
    \includegraphics[width=\linewidth]{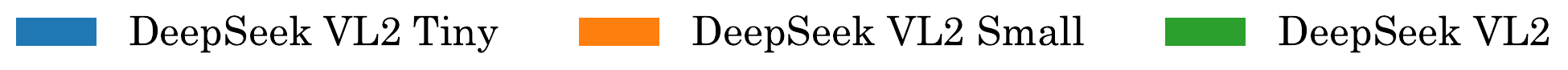}
    \begin{subfigure}{0.3\linewidth}
        \centering
        \includegraphics[width=\linewidth]{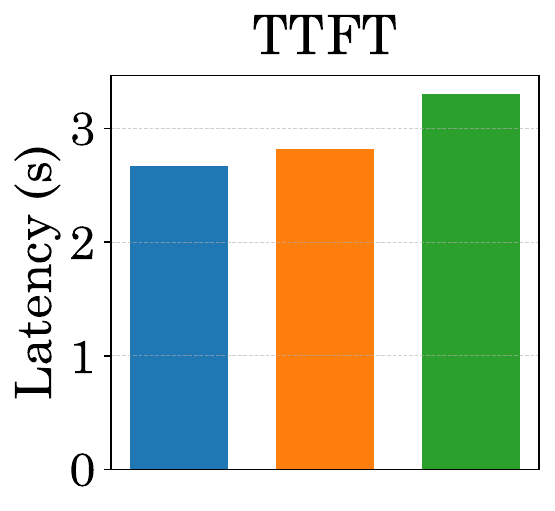}
    \end{subfigure}
    \hfill
    \begin{subfigure}{0.3\linewidth}
        \centering
        \includegraphics[width=\linewidth]{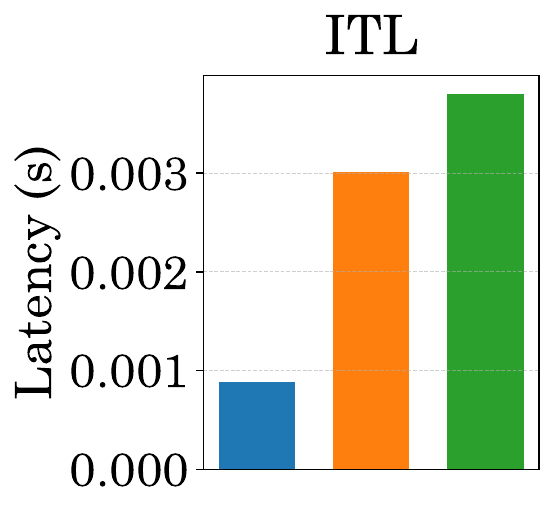}
    \end{subfigure}
    \hfill
    \begin{subfigure}{0.3\linewidth}
        \centering
        \includegraphics[width=\linewidth]{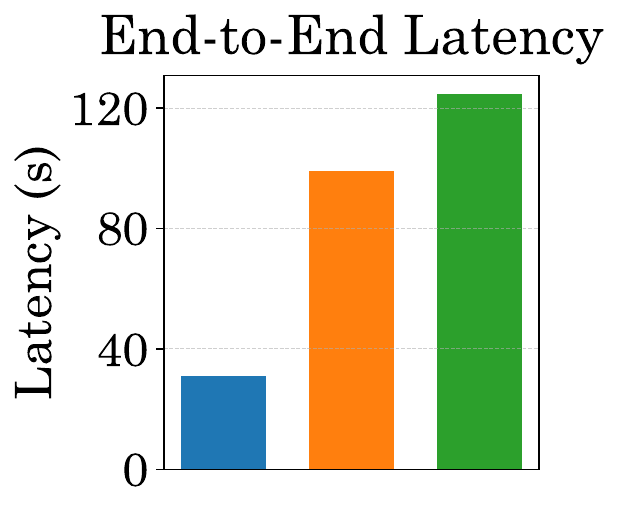}
    \end{subfigure}
    \vspace{-5mm} 
    \caption{TTFT, ITL and End-to-End Latency of VLMs} 
    \label{fig:VLM_prefill_decode}
    \vspace{-1mm}
\end{figure}

Figures \ref{fig:LLM_prefill_decode} and \ref{fig:VLM_prefill_decode} present a comparative analysis of TTFT, ITL, and end-to-end latency across LLMs and VLMs. Among LLMs, OLMoE-1B-7B achieves the fastest TTFT, outperforming DeepSeek-V2-Lite by approximately 70\%, while ITL varies by nearly 100\% between the best and worst performing models, and end-to-end latency shows over a 120\% gap. In VLMs, latency differences are more pronounced as DeepSeek-VL2-Tiny attains a TTFT about 30\% faster than the DeepSeek-VL2 model, with ITL showing a 240\% gap and end-to-end latency exceeding a 260\% difference. These results indicate that while LLMs exhibit moderate variation in latency, VLMs incur substantially larger performance gaps, mainly due to heavier computational load and multimodal processing overhead in the vision language inference pipeline.\looseness=-1


\subsection{Impact of Batch Size} 
\begin{figure}
    \centering
     \includegraphics[width=0.7\linewidth]{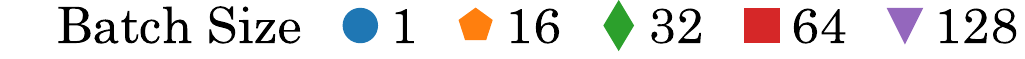}
    \begin{subfigure}{0.49\linewidth}
        \centering
        \includegraphics[width=0.9\linewidth]{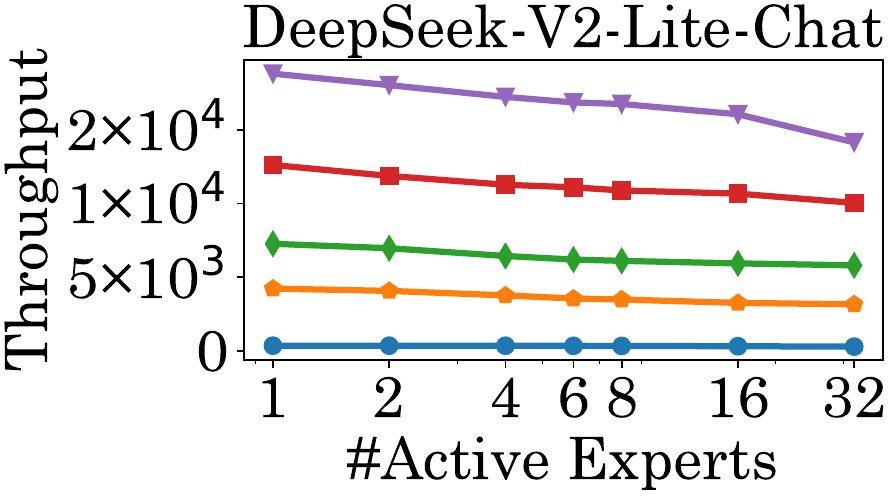}
    \end{subfigure}
    \begin{subfigure}{0.49\linewidth}
        \centering
        \includegraphics[width=0.9\linewidth]{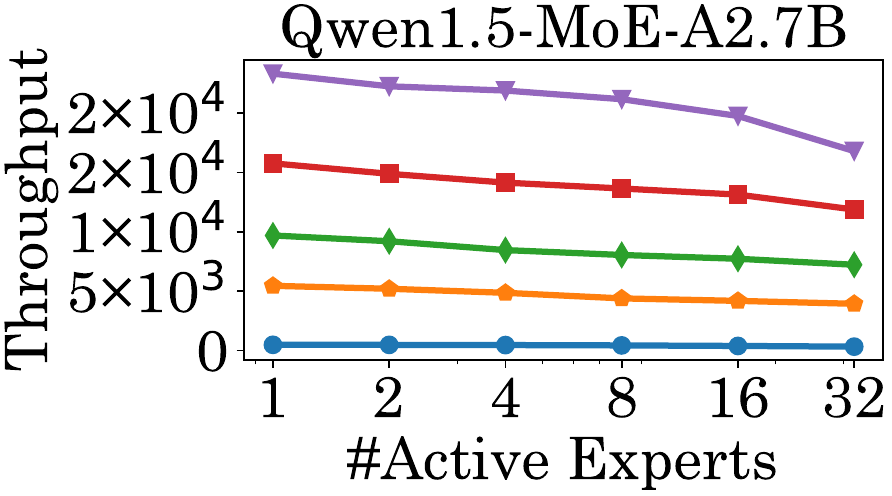}
    \end{subfigure}
    \vspace{-2mm}
    \caption{Impact of Batch Size on Varying Number of Active Experts (TopK) on Nvidia H100 GPU for Context Length 2048}
    \label{fig:topk_profiling}
\end{figure}

LLMs and VLMs demonstrate an increase in throughput with an increase in batch size for the same context length. This is mainly due to the simultaneous execution of all input prompts and the parallel generation of output tokens of all batches. 
Figure \ref{fig:topk_profiling} demonstrates that throughput consistently decreases as the number of active experts increases across DeepSeek-V2-Lite and Qwen1.5-MoE-A2.7B models and all batch sizes, with the performance degradation being more pronounced at higher batch sizes. For DeepSeek-V2-Lite, increasing the active experts from 1 to 32 results in an average throughput drop of approximately 15-20\% for large batch sizes (64 and 128), whereas small batch sizes (1 and 16) incur only about a 5-8\% reduction. Qwen1.5-MoE-A2.7B exhibits a similar trend but with slightly lower relative losses (around 12-18\% for large batches and 4-7\% for small batches), indicating a marginally better resilience to TopK scaling. Across both models, throughput scales sub-linearly with batch size: moving from batch size 1 to 128 increases throughput by roughly two orders of magnitude, but larger batches increase the throughput with more active experts. \looseness=-1 

\textit{\textbf{Insight}}: The results suggest that while larger batch sizes maximize hardware utilization, they are more sensitive to increased expert activation, highlighting a critical trade-off with batching and active experts in MoE inference.\looseness=-1

\subsection{Impact of Input/Output Sizes}

\begin{figure}[H]
\centering
    \includegraphics[width=\linewidth]{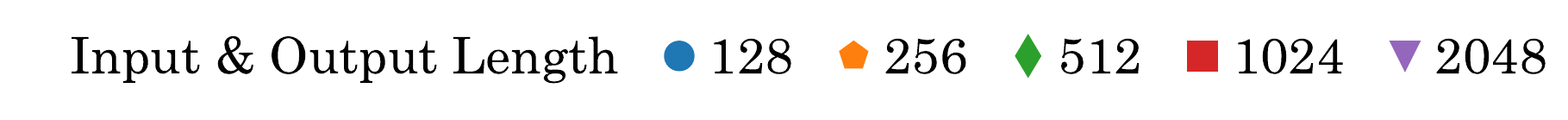}
    \begin{subfigure}{0.45\linewidth}
        \centering
        \includegraphics[width=\linewidth]{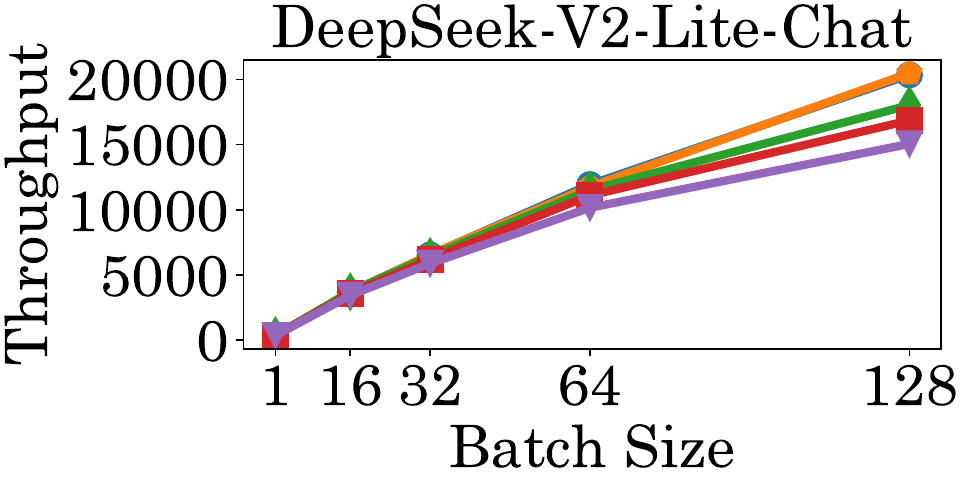}
    \end{subfigure}
    \begin{subfigure}{0.45\linewidth}
        \centering
        \includegraphics[width=\linewidth]{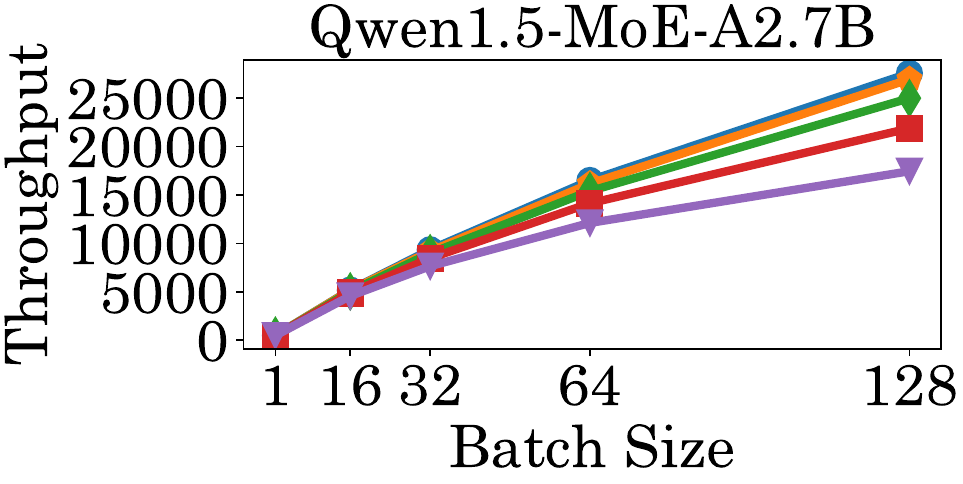}
    \end{subfigure}
    \caption{Batch Size vs Input \& Output Length on H100 GPU}
    \label{fig:batch_size_vs_input_output}
\end{figure}

Figure \ref{fig:batch_size_vs_input_output} illustrates throughput trends across varying batch sizes and sequence lengths for DeepSeek-V2-Lite and Qwen1.5-MoE-A2.7B. For both models, throughput scales linearly with batch size, with increases exceeding 8$\times$ from batch~1 to~128. Shorter sequences consistently outperform longer ones. For example, in both models, input/output length of 128 achieves up to 30\% higher throughput than length of 2048 at large batches, and the performance gap between shortest and longest sequences widens with batch size due to reduced memory and compute demands. For DeepSeek-V2-Lite, intermediate lengths (256-512) show less than 10\% drop compared to the shortest length, indicating efficient handling of medium sequences, whereas longer sequences (1024-2048) experience throughput degradation exceeding 20\%. Qwen1.5-MoE-A2.7B not only surpasses DeepSeek-V2-Lite by 20--30\% across all settings but also exhibits a more gradual decline in throughput as sequence length increases, reflecting better optimization for long-context inference.\looseness=-1

\textbf{\textit{Insight:}}
Throughput decreases as the output generation tokens increase. This is due to the increase in sequential token generation. Also, throughput increases with an increase in input length, as there is less opportunity for parallel processing. This reflects the fundamental difference between parallel input encoding and sequential output generation in transformer architectures. \looseness=-1

\section{Fine Grained MoE Inference Analysis}

\begin{figure*}
    \centering
    \includegraphics[width=0.75\textwidth]{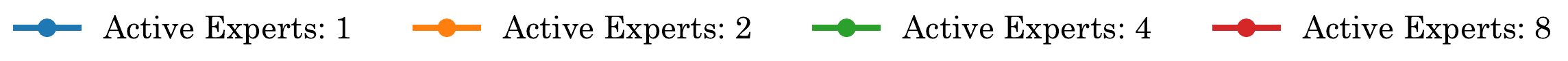}\\[0.2ex]
    \begin{subfigure}[b]{0.24375\textwidth}
        \centering
        \includegraphics[width=\linewidth]{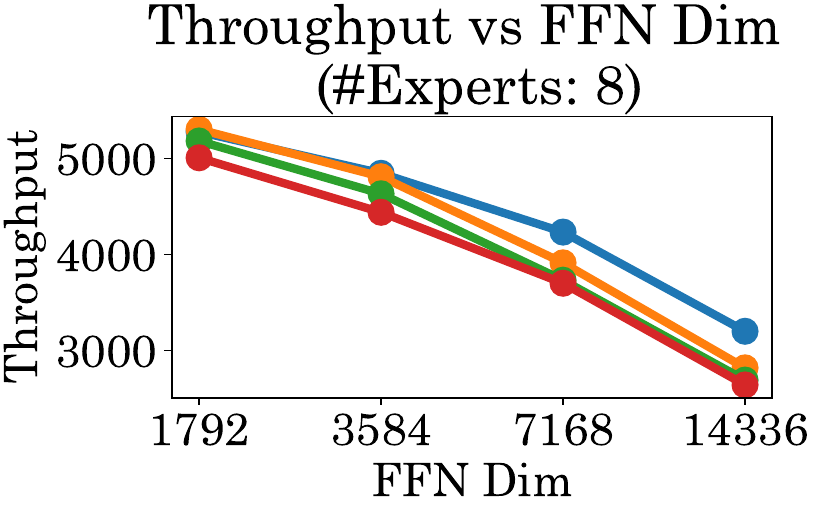}
        \caption{\centering \#Experts = 8}
    \end{subfigure}\hspace{-0.7pt}
    \begin{subfigure}[b]{0.24375\textwidth}
        \centering
        \includegraphics[width=\linewidth]{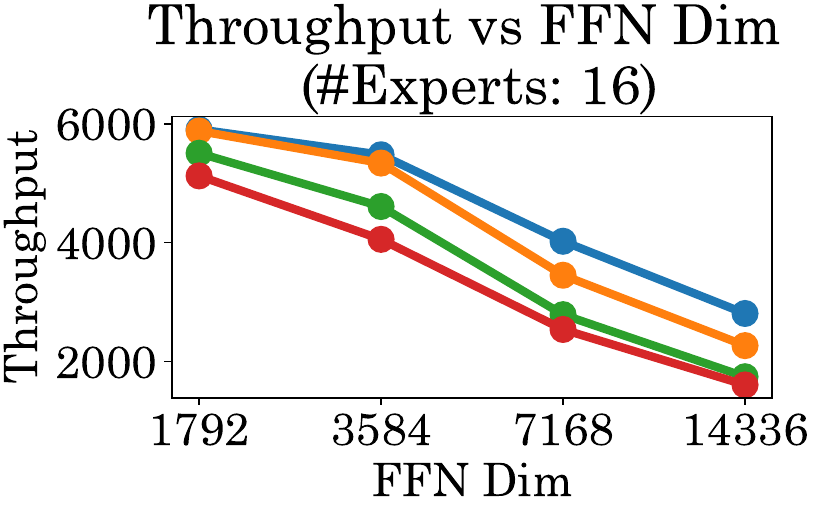}
        \caption{\centering \#Experts = 16}
    \end{subfigure}\hspace{-0.7pt}
    \begin{subfigure}[b]{0.24375\textwidth}
        \centering
        \includegraphics[width=\linewidth]{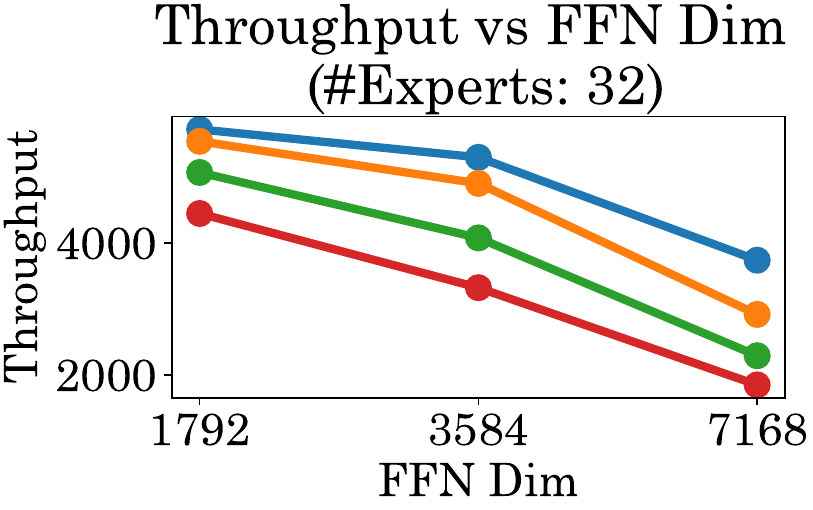}
         \caption{\centering \#Experts = 32}
    \end{subfigure}\hspace{-0.7pt}
    \begin{subfigure}[b]{0.24375\textwidth}
        \centering
        \includegraphics[width=\linewidth]{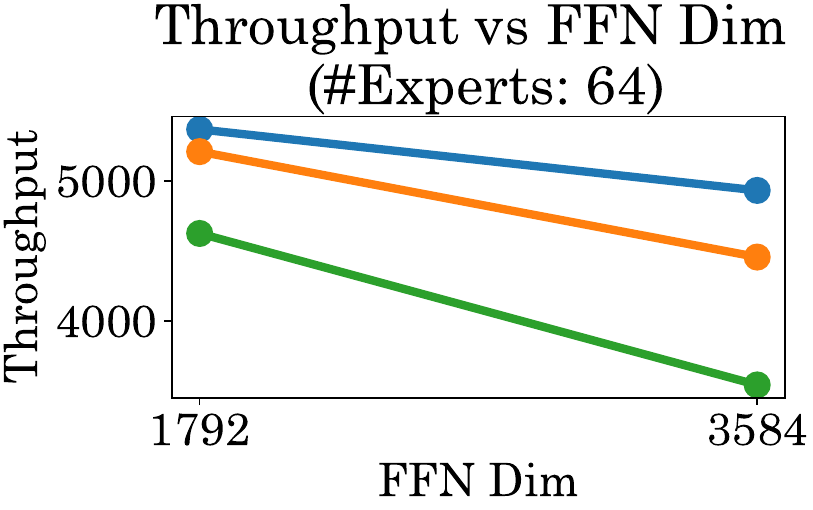}
        \caption{\centering \#Experts = 64}
    \end{subfigure}\hspace{-0.2pt}
    \vspace{-3mm}
    \caption{Throughput vs. FFN Dimension for Batch Size 16 and Input/Output Length 2048 on 4 H100 GPUs on Mixtral-8x7B Variant}
    \label{fig:fine_grained_moe_1}
\end{figure*}
\begin{figure*}[t]
    \centering
    \includegraphics[width=0.75\textwidth]{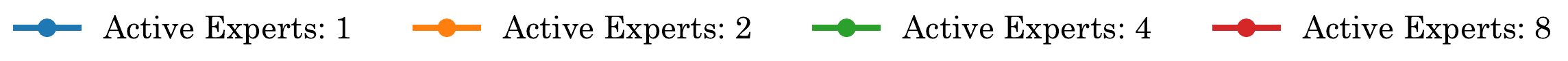}\\[0.2ex]
    \begin{subfigure}[b]{0.24375\textwidth}
        \centering
        \includegraphics[width=\linewidth]{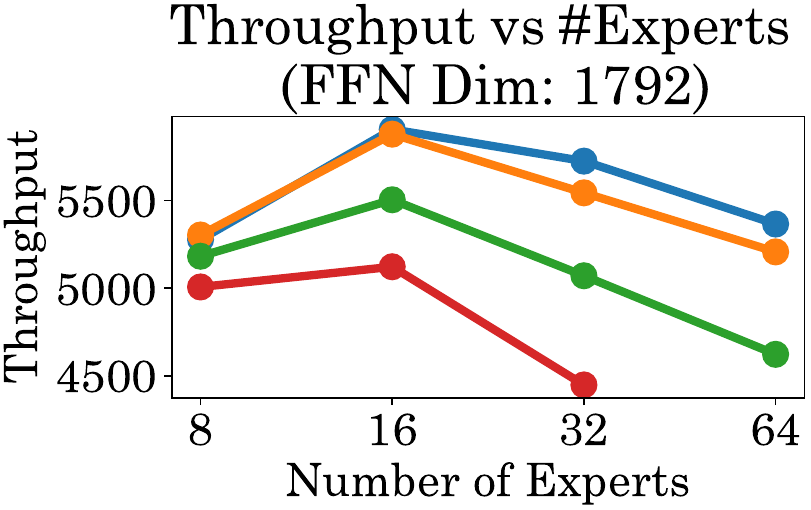}
        \caption{\centering FFN Dim = 1792}
    \end{subfigure}\hspace{-0.7pt}
    \begin{subfigure}[b]{0.24375\textwidth}
        \centering
        \includegraphics[width=\linewidth]{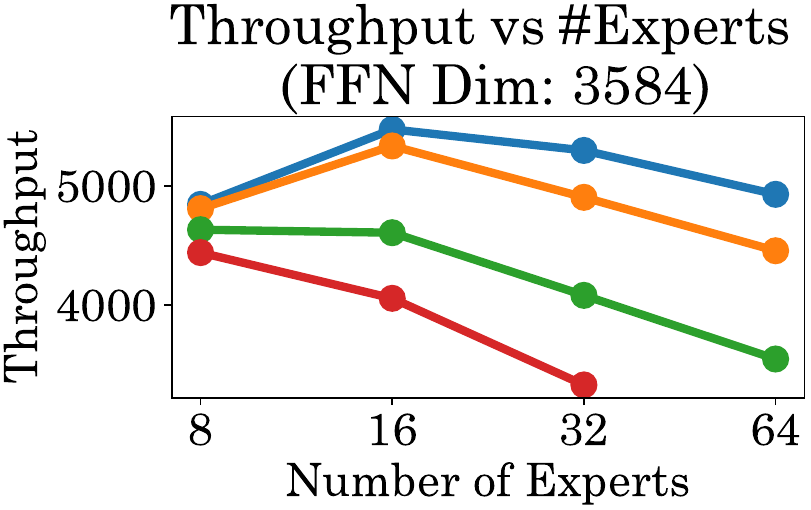}
        \caption{\centering FFN Dim = 3584}
    \end{subfigure}\hspace{-0.7pt}
    \begin{subfigure}[b]{0.24375\textwidth}
        \centering
        \includegraphics[width=\linewidth]{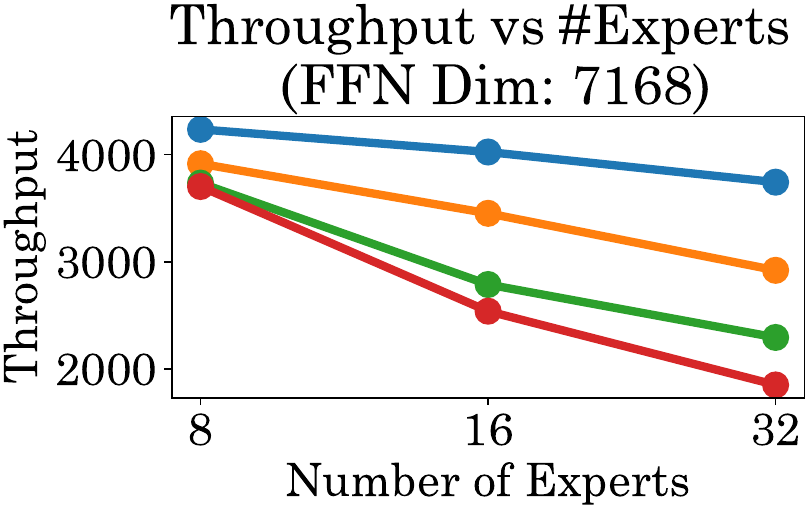}
         \caption{\centering FFN Dim = 7168}
    \end{subfigure}\hspace{-0.7pt}
    \begin{subfigure}[b]{0.24375\textwidth}
        \centering
        \includegraphics[width=\linewidth]{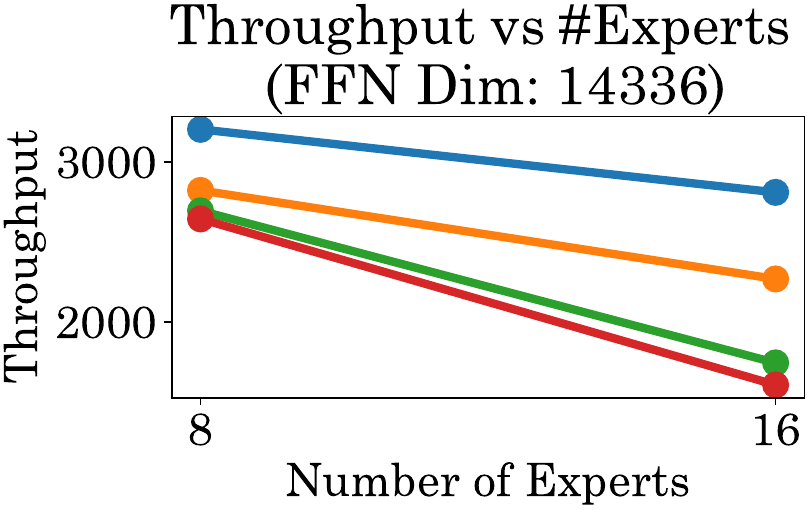}
        \caption{\centering FFN Dim = 14336}
    \end{subfigure}\hspace{-0.2pt}
    \vspace{-3mm}
    \caption{Throughput vs. \#Experts for Batch Size 16 and Input/Output Length 2048 on 4 H100 GPUs on Mixtral-8x7B Variant}
    \label{fig:fine_grained_moe_2}
\end{figure*}
\begin{figure*}[t]
    \centering
    \includegraphics[width=0.75\textwidth]{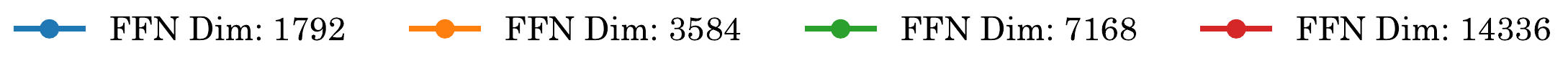}\\[0.2ex]
    \begin{subfigure}[b]{0.24375\textwidth}
        \centering
        \includegraphics[width=\linewidth]{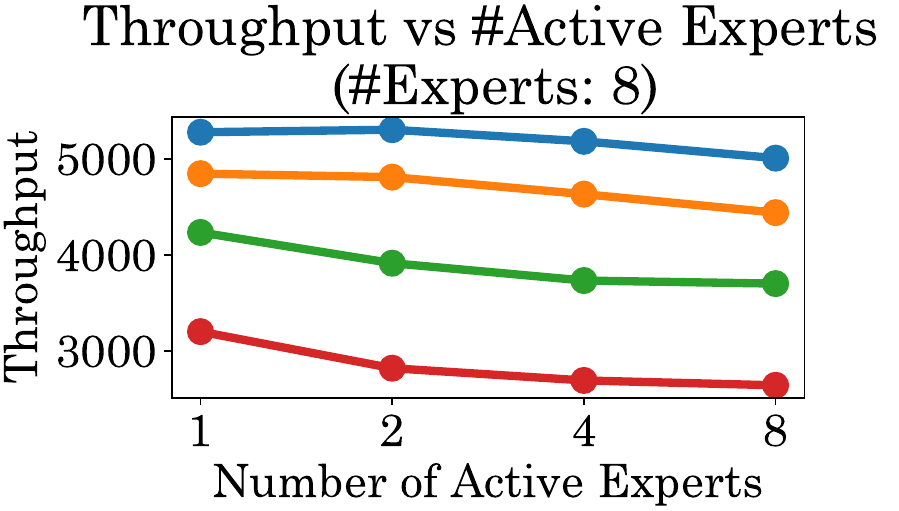}
        \caption{\centering \#Experts = 8}
    \end{subfigure}\hspace{-0.7pt}
    \begin{subfigure}[b]{0.24375\textwidth}
        \centering
        \includegraphics[width=\linewidth]{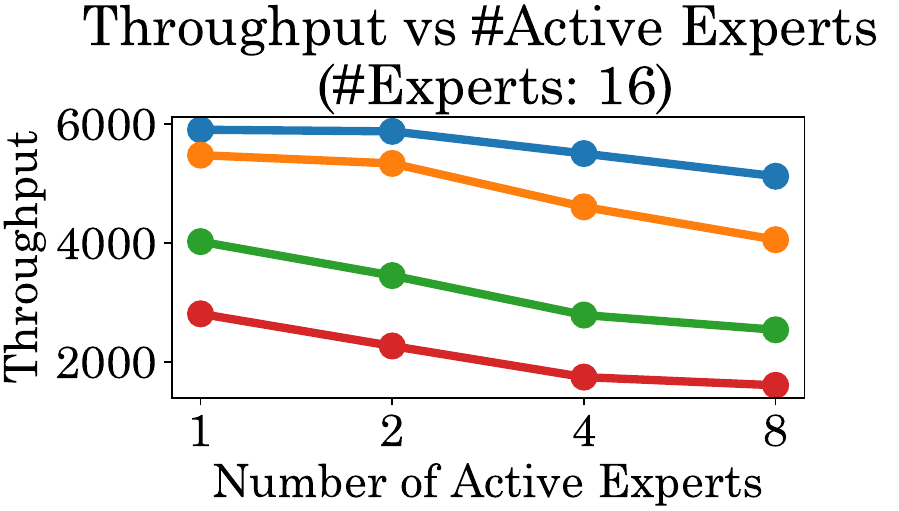}
        \caption{\centering \#Experts = 16}
    \end{subfigure}\hspace{-0.7pt}
    \begin{subfigure}[b]{0.24375\textwidth}
        \centering
        \includegraphics[width=\linewidth]{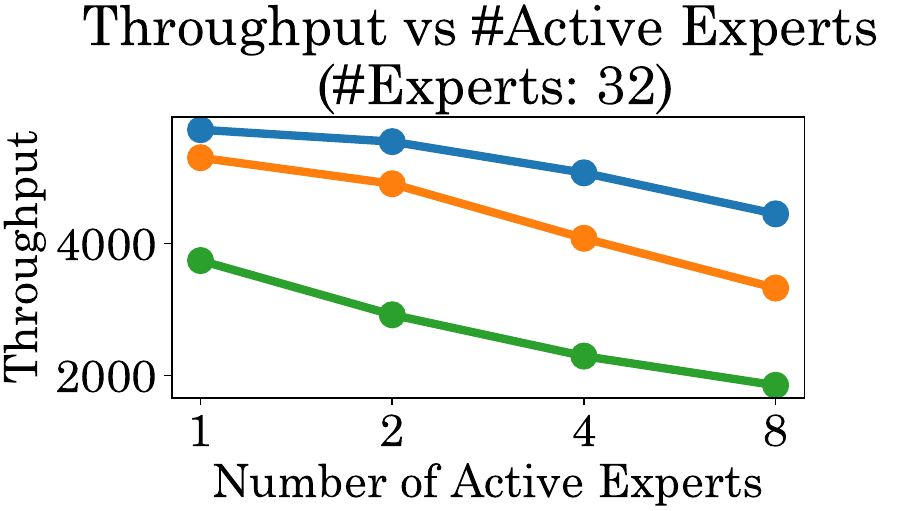}
         \caption{\centering \#Experts = 32}
    \end{subfigure}\hspace{-0.7pt}
    \begin{subfigure}[b]{0.24375\textwidth}
        \centering
        \includegraphics[width=\linewidth]{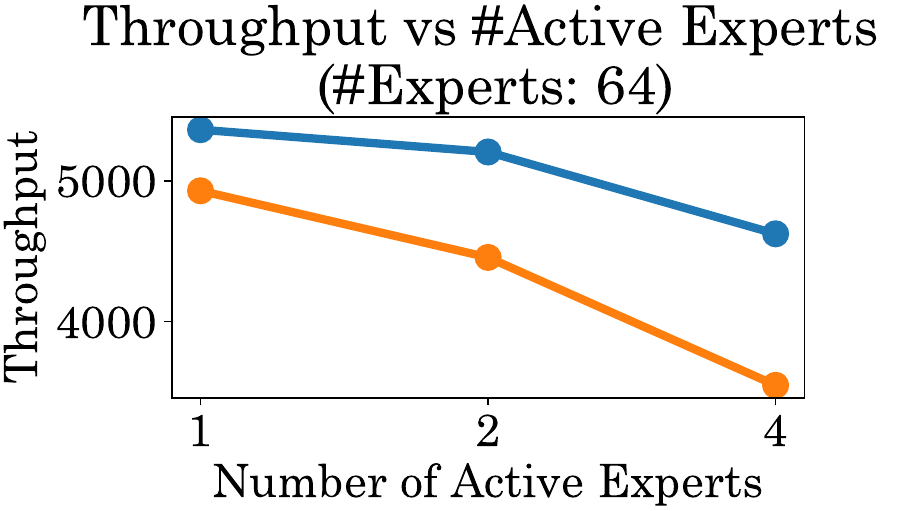}
        \caption{\centering \#Experts = 64}
    \end{subfigure}\hspace{-0.2pt}
    \vspace{-3mm}
    \caption{Throughput vs. \#Active Experts for Batch Size 16 and Input/Output Length 2048 on 4 H100 GPUs on Mixtral-8x7B Variant}
    \label{fig:fine_grained_moe_3}
\end{figure*}

\subsection{Hyperparameter Setup}
This section investigates the impact of scaling MoE hyperparameters in a layer. We explore several possible combinations within our predefined hyperparameter configuration, namely FFN dimension: $\{1792, 3584, 7168, 14336\}$, number of experts: $\{8, 16, 32, 64\}$, and number of active experts: $\{1, 2, 4, 8\}$. The baseline skeleton model is Mixtral-8x7B and we tweak the hyperparameters in each experiment. All experiments are conducted on 4 H100 GPUs using vLLM. Any missing data points in the results indicate OOM conditions.

\subsection{Scaling FFN Dimension}

Figure \ref{fig:fine_grained_moe_1} illustrates the scaling of FFN dimension for a fixed number of experts.  Across all expert configurations, throughput steeply declines by 50\% on average when FFN dimension increases from 1792 to 14336, with the steepest drops occurring in the transition from 1792 to 3584. This performance degradation is particularly acute for configurations with higher active expert counts, where 8 active expert scenarios consistently show the most throughput reductions. The impact of active experts becomes increasingly impactful at higher FFN dimensions, with single active expert configurations maintaining relatively stable throughput compared to multiple active expert scenarios. At the largest FFN dimension (14336), the performance gap between one active and eight active expert configurations reaches around 60\%, highlighting the effect of increased data movement and computation overhead. The asymptotic behavior observed at the highest FFN dimensions across all configurations suggests approaching the theoretical bandwidth limits of H100. 

\textbf{\textit{Insight:}} The convergence of throughput, regardless of active expert count at extreme FFN sizes, indicates that memory bandwidth saturation overrides computational parallelism benefits. This finding has critical implications for MoE deployment strategies, suggesting that practitioners should carefully balance FFN capacity against throughput requirements.

\subsection{Scaling Number of Experts}
Figure \ref{fig:fine_grained_moe_2} illustrates the scaling of the number of experts for a fixed FFN dimension. 
The scaling patterns with total expert count show a complex non-linear relationship that varies significantly based on FFN dimension and active experts. For smaller FFN dimensions (1792, 3584), increasing the total number of experts from 8 to 64 generally maintains or slightly improves throughput, with improvements ranging from 5-15\% in optimal configurations. However, this positive scaling behavior becomes increasingly constrained at larger FFN dimensions, where the additional expert capacity cannot be effectively utilized due to memory bandwidth limitations. The interaction between total experts and active experts shows a resource allocation challenge that becomes more complex with increasing scale. Configurations with higher active expert counts (4, 8) show diminishing returns more rapidly as total experts increase, particularly evident in the flattening throughput curves beyond 32 total experts. 

\textbf{\textit{Insight:} }
As number of experts grow, routing and communication overhead can overshadow computational gains, while memory limits, especially in high FFN configurations, cause out-of-memory failures. Effective MoE deployment should optimize the total parameter budget rather than maximize expert count, with extreme scale configurations likely needing distributed placement across multi-node architectures for efficient resource use.

\subsection{Scaling Number of Active Experts}

Figure \ref{fig:fine_grained_moe_3} illustrates the scaling of the number of active experts for a fixed FFN dimension. The active expert scaling reveals a consistent throughput degradation as the number of active experts increases from 1 to 8 across all configurations. Single active expert configurations consistently deliver 50-80\% higher throughput compared to 8 active expert scenarios, representing an efficiency optimization opportunity in MoE deployment strategies. This substantial performance difference reflects the fundamental relationship between sparse activation benefits and multi-expert overhead, particularly evident in the linear throughput degradation patterns observed across different total expert and FFN configurations. The consistency of this degradation across varying total expert counts suggests that active expert management represents a primary optimization level for inference production deployments. The scaling behavior across FFN dimensions reveals that active expert overhead is not uniformly distributed across different settings. At smaller FFN dimensions, the throughput gap between 1 active and 8 active configurations remains relatively modest (20-30\%), while at larger FFN dimensions this gap expands dramatically (60-80\%). The interaction suggests that high-capacity MoE configurations may benefit from dynamic active expert allocation strategies that adjust based on computation and memory availability.

\textbf{\textit{Insight:}} MoE throughput drops sharply with more active experts, with single expert setups delivering up to 80\% higher performance at larger FFN sizes. Jointly tuning expert count, FFN dimension, and activation strategy is essential, as smaller FFNs allow flexibility while larger ones require conservative activation to avoid OOM. 

To summarize our findings on scaling the number of experts, active experts, and FFN dimensions , the data reveals clear operating regimes where different parameter combinations provide optimal throughput characteristics, with smaller FFN dimensions (1792-3584) enabling more flexible active expert usage while larger dimensions (7168-14336) require more conservative activation strategies to maintain acceptable throughput. The systematic OOM boundaries observed at extreme configurations provide deployment guidelines for hardware-constrained environments, indicating that current H100-based systems can effectively support MoE models up to specific parameter budgets before requiring distributed architectures. 

\section{MoE Algorithm Optimizations}

\subsection{Quantization} 
\begin{figure}[H]
    \centering
    \begin{subfigure}{0.48\linewidth}
        \centering
        \includegraphics[width=\linewidth]{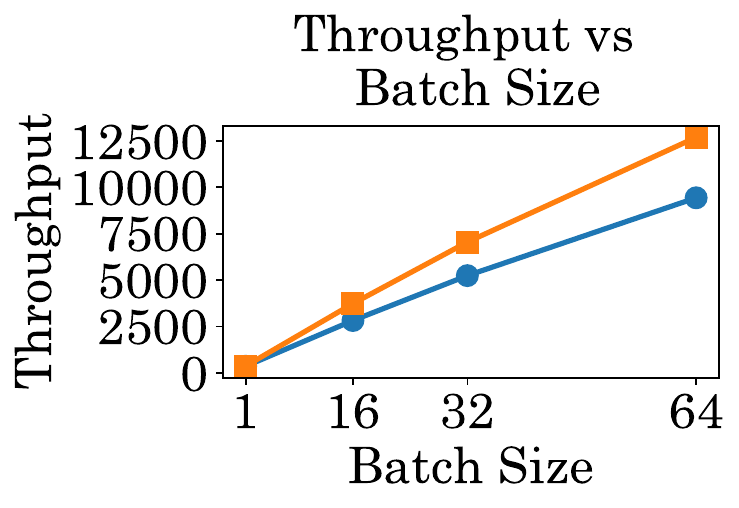}
    \end{subfigure}
    \hfill
    \begin{subfigure}{0.48\linewidth}
        \centering
        \includegraphics[width=\linewidth]{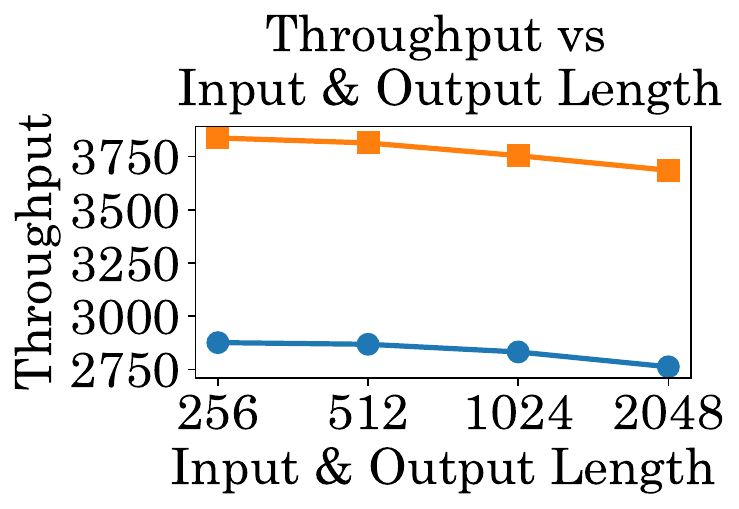}
    \end{subfigure}
    \includegraphics[width=0.4\linewidth]{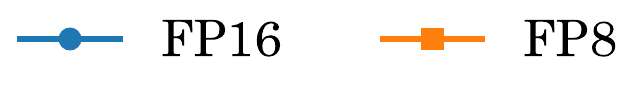}
    \vspace{-1mm} 
    \caption{Performance Comparison of Mixtral-8x7B with FP16 and FP8 precisions on Nvidia H100 GPUs}
    \label{fig:quantization}
\end{figure}

Quantization \cite{gholami2022survey} is a method to reduce model size by lowering the precision of weights and activations. 
LLMs can be operated in lower precisions, such as FP8 \cite{kuzmin2022fp8}, using GPTQ \cite{frantar2022gptq} and AWQ \cite{lin2023awq} without compromising the model quality. Figure \ref{fig:quantization} compares the performance of Mixtral-8x7B under FP16 and FP8 precisions using vLLM on H100 GPU with varying batch sizes and input/output lengths. Across both settings, FP8 outperforms FP16 in throughput, with the performance gap widening under larger batch sizes and remaining stable across varying sequence lengths. Specifically, FP8 achieves up to ~25–30\% higher throughput than FP16 at the highest batch size, indicating superior scalability with parallel workloads. In the input/output length variation analysis, FP8 sustains a throughput advantage of around 20–25\% over FP16 across all tested lengths, suggesting that the benefit of lower precision is robust to changes in sequence length and not limited to small context inference. 

\textit{\textbf{Insight:}} These results show FP8’s potential to deliver substantial efficiency gains in both compute-bound and memory-bound scenarios on H100 GPUs.



\subsection{\textbf{MoE Pruning}}

\begin{figure}
    \centering
    \begin{subfigure}{0.48\linewidth}
        \centering
        \includegraphics[width=\linewidth]{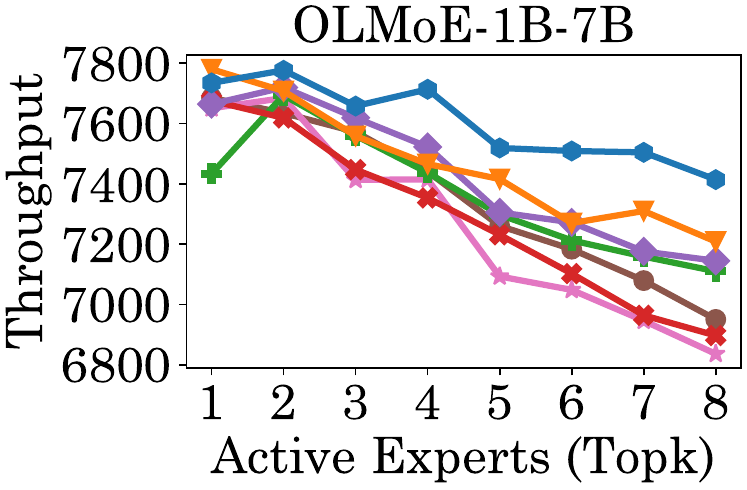}
    \end{subfigure}
    \hfill
    \begin{subfigure}{0.48\linewidth}
        \centering
        \includegraphics[width=\linewidth]{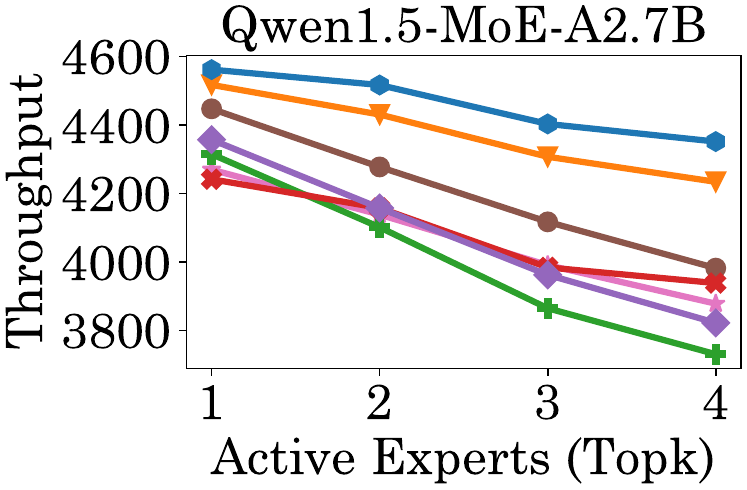}
    \end{subfigure}
    \includegraphics[width=\linewidth]{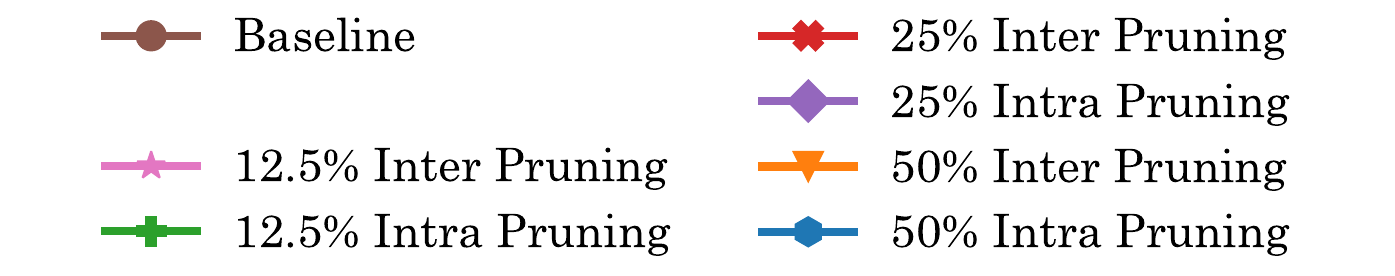}
    \vspace{-5mm} 
    \caption{Impact of Intra and Inter Expert Pruning on 4 H100 GPUs for Batch Size 16 and Input/output Length of 2048}
    \label{fig:pruning}
    \vspace{-5mm}
\end{figure}

Inter-expert pruning \cite{lu2024not} removes an entire expert along with their routing weights, reducing memory while keeping the same number of active experts during inference. Intra-expert pruning \cite{yang2024moe} reduces the FFN Dimension inside each expert, keeping the number of experts unchanged but lowering the computation per expert. In our experiments, we apply pruning ratios of \{12.5\%, 25\%, 50\%\}. For example, 12.5\% inter-expert pruning removes $1/8$ of the experts in each layer, while 25\% intra-expert pruning reduces the FFN dimension by 1/4. We evaluate TopK values from 1 up to the baseline pretrained top-$k$: $\{1, 2, \ldots, \text{TopK}_{\text{baseline}}\}$. The results in Figure~\ref{fig:pruning} show that throughput generally decreases as the number of active experts increases, with intra- and inter-expert pruning exhibiting distinct trends across models. For OLMoE-1B-7B, higher pruning ratios (e.g., 50\%), particularly intra-expert pruning tend to sustain or even improve throughput for larger TopK, likely due to reduced per-expert computation enabling better hardware utilization. In contrast, Qwen1.5-MoE-A2.7B is more sensitive to pruning, where aggressive intra-expert pruning at low TopK significantly degrades throughput, indicating greater vulnerability to load imbalance. On NVIDIA H100 GPUs, these effects are amplified because the GPU’s high compute-to-memory bandwidth ratio and advanced scheduling mechanisms make performance more sensitive to expert load balancing; when token-to-expert routing is imbalanced, some experts become bottlenecks, reducing the overall parallel efficiency despite the available compute capacity. Low pruning percentages (12.5\% or 25\%) of inter and intra expert pruning can cause an inverse effect and reduce throughput, while 50\% pruning can significantly improve throughput.

\subsection{Speculative Decoding Study}

Speculative decoding is a technique to accelerate LLM inference by generating multiple tokens in parallel and verifying them. The process involves a small and lightweight draft model that generates several future tokens in a single step, followed by a verification step using the larger, more accurate model to validate or reject the sequence. This approach reduces the number of sequential forward passes required, significantly improving decoding throughput while maintaining output quality. Recent implementations integrate speculative decoding with advanced scheduling and KV cache management, making it particularly effective for real-time and large-scale deployment scenarios. A key limitation of speculative decoding is that the main model and the draft model must share an identical vocabulary. Consequently, the two models are typically selected from the same family, Qwen, to ensure compatibility.

\begin{figure}
    \centering
    \begin{subfigure}{0.48\linewidth}
        \centering
        \includegraphics[width=\linewidth]{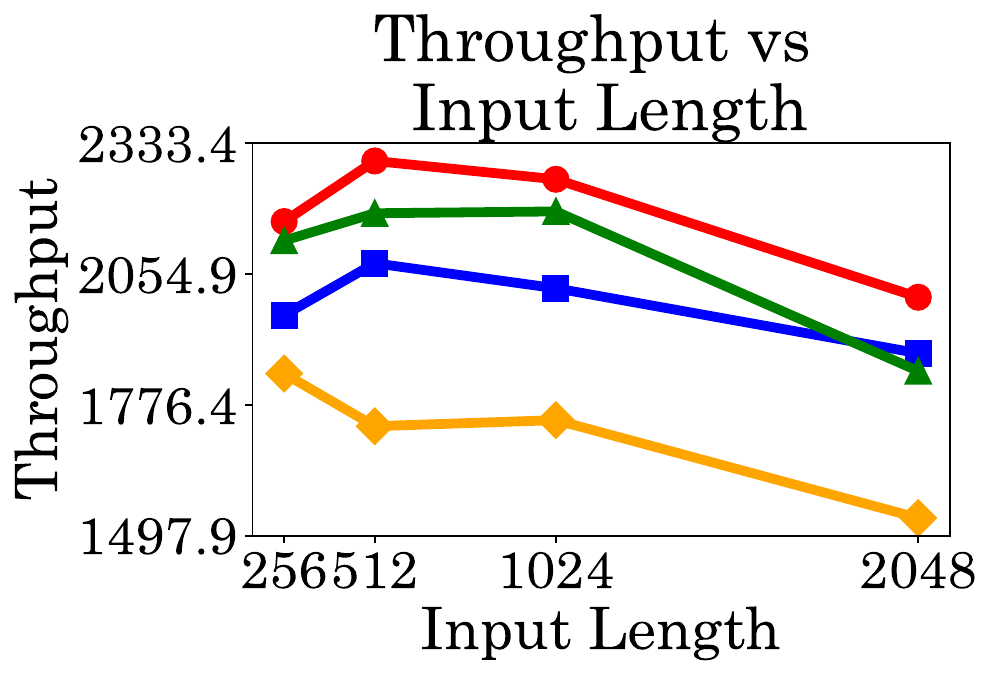}
    \end{subfigure}
    \hfill
    \begin{subfigure}{0.48\linewidth}
        \centering
        \includegraphics[width=\linewidth]{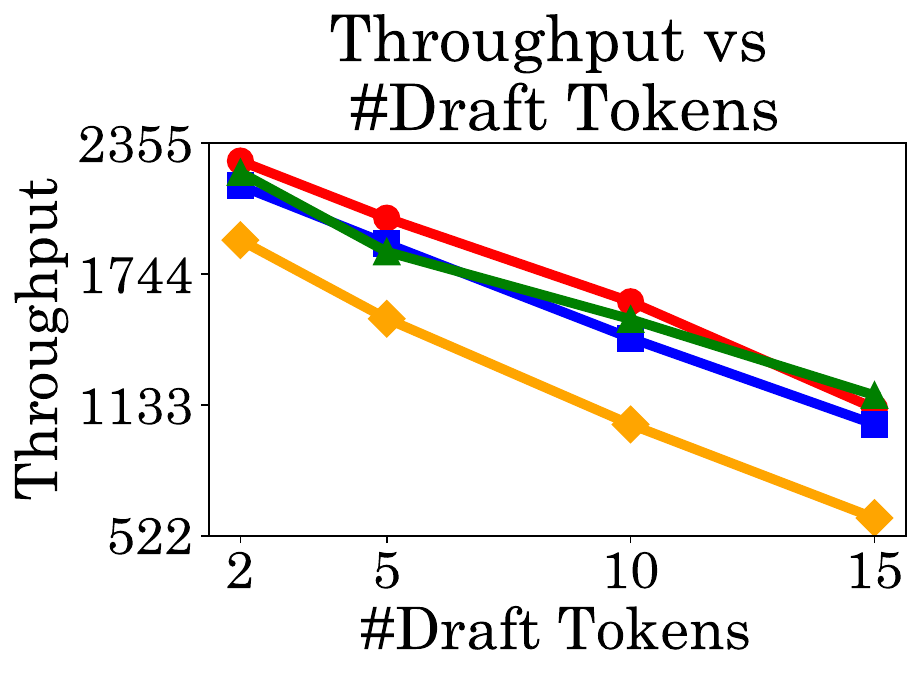}
    \end{subfigure}
     \includegraphics[width=0.6\linewidth]{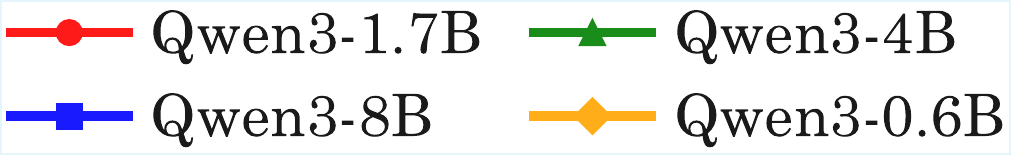}
    \vspace{-2mm} 
    \caption{Comparison of Speculative Decoding Performance on Target Model Qwen3-30B-A3B using four Draft Models}
    \label{fig:spec_decoding}
    \vspace{-5mm}
\end{figure}

Figure \ref{fig:spec_decoding} compares the speculative decoding performance of Qwen-30B using four draft models from the same family, Qwen3-0.6B, Qwen3-1.7B, Qwen3-4B and Qwen3-8B. Qwen-30B as the target model shows that Qwen3-1.7B delivers the highest throughput, exceeding Qwen3-8B by up to $\sim$20\% at short inputs and retaining a $\sim$15\% lead over Qwen3-4B at long inputs, while Qwen3-0.6B lags by $\sim$25-35\% across all lengths. Throughput drops with increasing input length for all models, but the decline is smaller ($\sim$15\%) for Qwen3-1.7B compared to $\sim$25\% for Qwen3-8B and Qwen3-4B, indicating better scalability. As draft tokens increase, throughput decreases monotonically due to higher validation overhead, with Qwen3-1.7B maintaining a $\sim$5-10\% advantage over Qwen3-4B and $\sim$10\% over Qwen3-8B at higher counts, while Qwen3-0.6B remains over $\sim$30\% slower than the leader. These trends highlight that medium-sized draft models balance accuracy and efficiency best, while very small or large drafts incur greater latencies.

\section{Hardware Optimizations}

\subsection{GPU Parallelism}


Tensor Parallelism (TP) \cite{shoeybi2019megatron} distributes layer weight tensors across multiple devices in either row-wise or column-wise fashion. Devices communicate to share input and output activations. TP works most effectively within single nodes due to faster intra-node communication, enabling the distribution of large tensors that exceed single-device memory capacity.
Expert Parallelism (EP) \cite{rajbhandari2022deepspeed} distributes MoE models by assigning groups of expert blocks to individual devices. This approach exploits the independent nature of experts in MoE layers, though it can suffer from load-balancing issues when assigned experts remain inactive.
Hybrid Parallelism (HP) \cite{singh2023hybrid} combines multiple parallelism strategies (TP, PP and EP) to achieve efficient scaling and improved hardware utilization. While HP provides greater flexibility by allowing different parallelism techniques per layer, it introduces complexity in managing simultaneous parallelism strategies and coordinating work distribution across devices.

\begin{figure}[H]
    \centering
    \begin{subfigure}{0.4\linewidth}
        \centering
        \includegraphics[width=\linewidth]{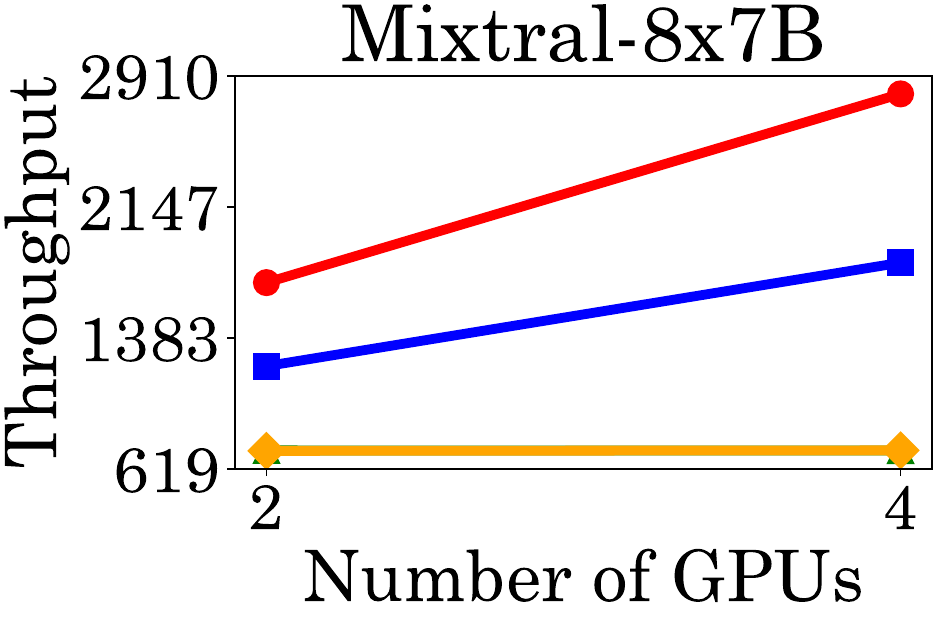}
    \end{subfigure}
    \hfill
    \begin{subfigure}{0.4\linewidth}
        \centering
        \includegraphics[width=\linewidth]{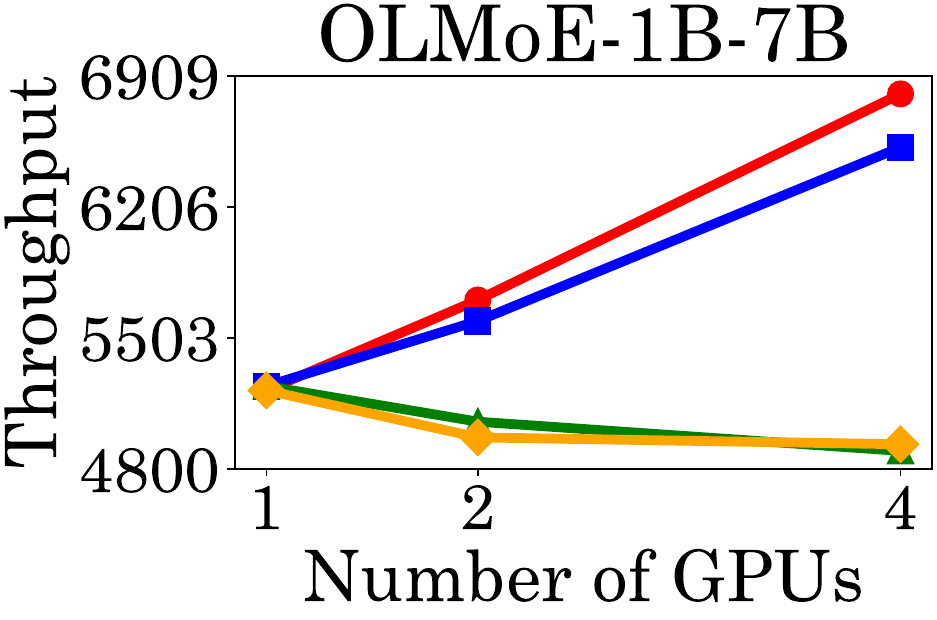}
    \end{subfigure}
    \includegraphics[width=0.5\linewidth]{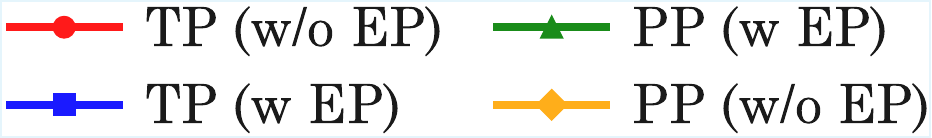}
    \vspace{-2mm} 
    \caption{Performance Comparison of Mixtral-8x7B using TP, PP, EP on Nvidia H100 GPUs using vLLM}
    \label{fig:TP_PP_EP}
\end{figure}

Figure \ref{fig:TP_PP_EP} illustrates the performance of the Mixtral-8x7B model and OLMoE-1B-7B models under different settings of TP, PP and EP. The results show that TP without EP delivers the highest throughput scaling as the number of GPUs increases, achieving performance gains of over $2\times$ from 1 to 4 GPUs on the H100. TP with EP exhibits lower scaling efficiency, while PP with EP shows minimal throughput improvement, and PP without EP remains almost flat, indicating poor scalability. This phenomenon on the H100 GPU arises because its high intra-node bandwidth (via NVLink) strongly benefits communication-intensive TP, allowing large weight tensors to be efficiently split and aggregated across devices. In contrast, PP suffers from stage imbalance and synchronization overheads, and EP’s load-balancing and dispatch costs offset potential gains, especially for smaller expert activations.

\textit{\textbf{Insight}:} Tensor parallelism over the entire model is more effective than pipeline or expert parallelism. This is due to better utilization of all available GPU devices, whereas expert and pipeline parallelism often result in underutilization of resources.





\subsection{Fused MoE}

\begin{figure}
    \centering
    \begin{subfigure}{0.48\linewidth}
        \centering
        \includegraphics[width=\linewidth]{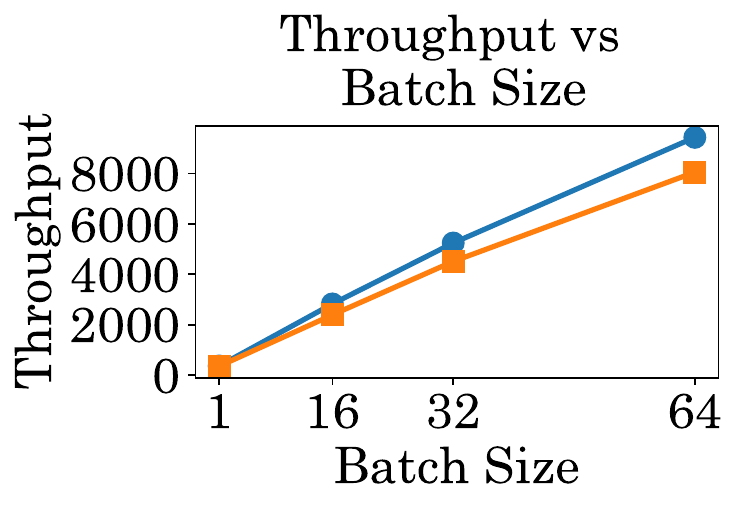}
    \end{subfigure}
    \hfill
    \begin{subfigure}{0.48\linewidth}
        \centering
        \includegraphics[width=\linewidth]{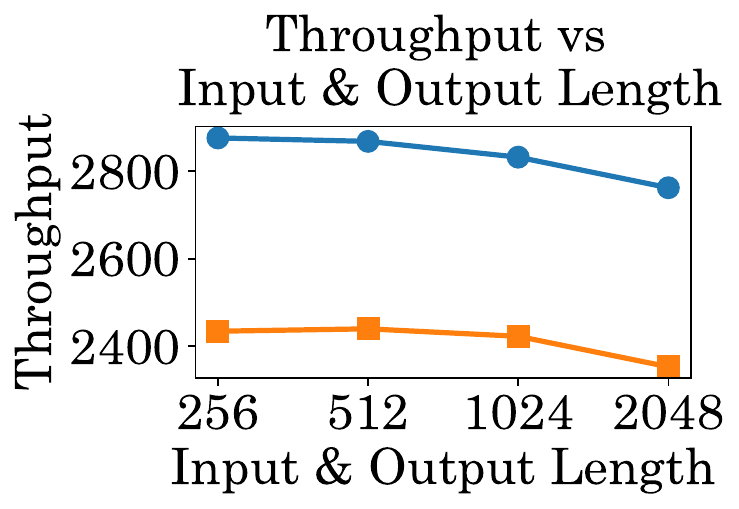}
    \end{subfigure}
    \includegraphics[width=\linewidth]{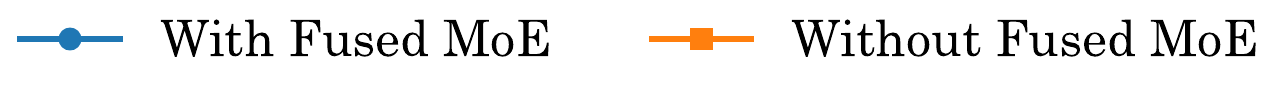}
    \vspace{-8mm} 
    \caption{Performance Comparison of Mixtral-8x7B with and without Fused MoE Configuration on 4 H100 GPUs}
    \label{fig:fused_moe}
    \vspace{-2mm}
\end{figure}

Fused MoE is an optimized execution for MoEs to merge expert selection, routing, and FFN computation into a single fused GPU kernel, reducing intermediate memory transfers and kernel launch overhead. Fused MoE minimizes synchronization costs and improves GPU utilization by batching token routing decisions and executing only the active experts in one pass, leading to significantly higher throughput compared to a naive MoE implementation where routing and expert computation are separate stages. Figure \ref{fig:fused_moe} shows the performance of Mixtral-8x7B with and without the Fused MoE, both varying batch size and input/output lengths. Across both settings, Fused MoE consistently outperforms the non-fused version, with performance gains becoming more pronounced at higher context lengths and prompts. When scaling batch size, Fused MoE achieves approximately 15–20\% higher throughput, with the relative advantage widening as the batch size increases, indicating superior GPU utilization and reduced kernel launch overhead. In the input/output length variation experiment, Fused MoE maintains a throughput advantage of roughly 12–18\% across all sequence lengths, while the non-fused baseline exhibits a sharper decline at longer sequences. 

\textit{\textbf{Insight}:} These results highlight that kernel fusion not only boosts throughput but also sustains efficiency under increasing computational and memory demands, aligning with its design goal of minimizing synchronization costs and intermediate memory transfers.\looseness=-1

\subsection{Hardware Benchmarking}

\begin{figure}
    \centering
    \includegraphics[width=0.6\linewidth]{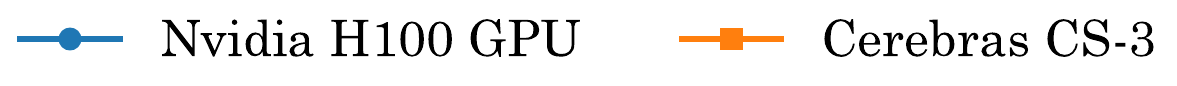}
    \begin{subfigure}{0.48\linewidth}
        \centering
        \includegraphics[width=\linewidth]{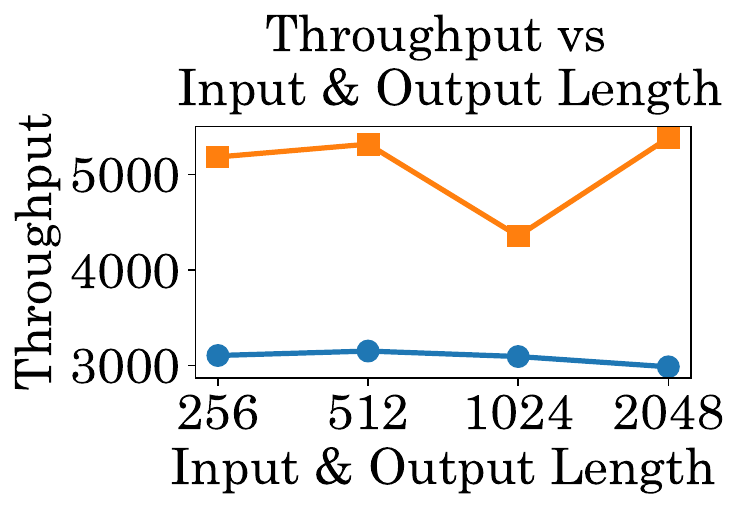}
    \end{subfigure}
    \hfill
    \begin{subfigure}{0.48\linewidth}
        \centering
        \includegraphics[width=\linewidth]{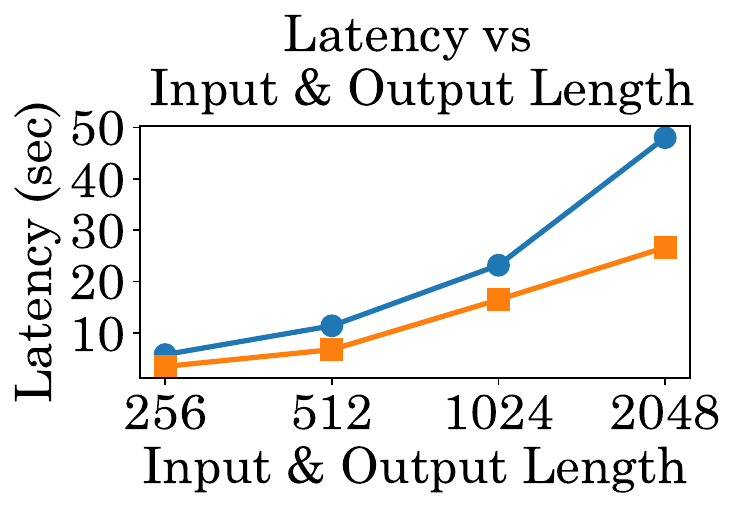}
    \end{subfigure}
    \vspace{-2mm}
    \caption{H100 vs CS3: Throughput and Latency Comparison of Llama-4-Scout-17B-16E}
    \label{fig:cs3_h100}
    \vspace{-5mm}
\end{figure}

Figure \ref{fig:cs3_h100} compares latency and throughput for Llama-4-Scout-17B-16E model on H100 GPU and Cerebras cloud CS-3 systems across varying input/output lengths. The CS-3 model replica stores most weights at FP8 precision, though KV cache and all computation are performed at FP16 for maximum accuracy. The latency increases more steeply on H100 with context length, with a sharp rise beyond 1024 tokens, while the CS-3 maintains significantly lower and more gradual latency growth, indicating better scalability. CS-3 benefits from WSE-3 having multiple orders of magnitude memory bandwidth and decreased inter-device communication, enabling rapid inference pipelining slowed only slightly by infrequent cross-node pipelining. We selected Llama-4 Scout as it is the only model with stable support across H100 and CS-3, enabling a fair comparison.

 \begin{figure*}
 \centering
        \subfloat[MolmoE-1B]{
            \includegraphics[width=.23\linewidth]{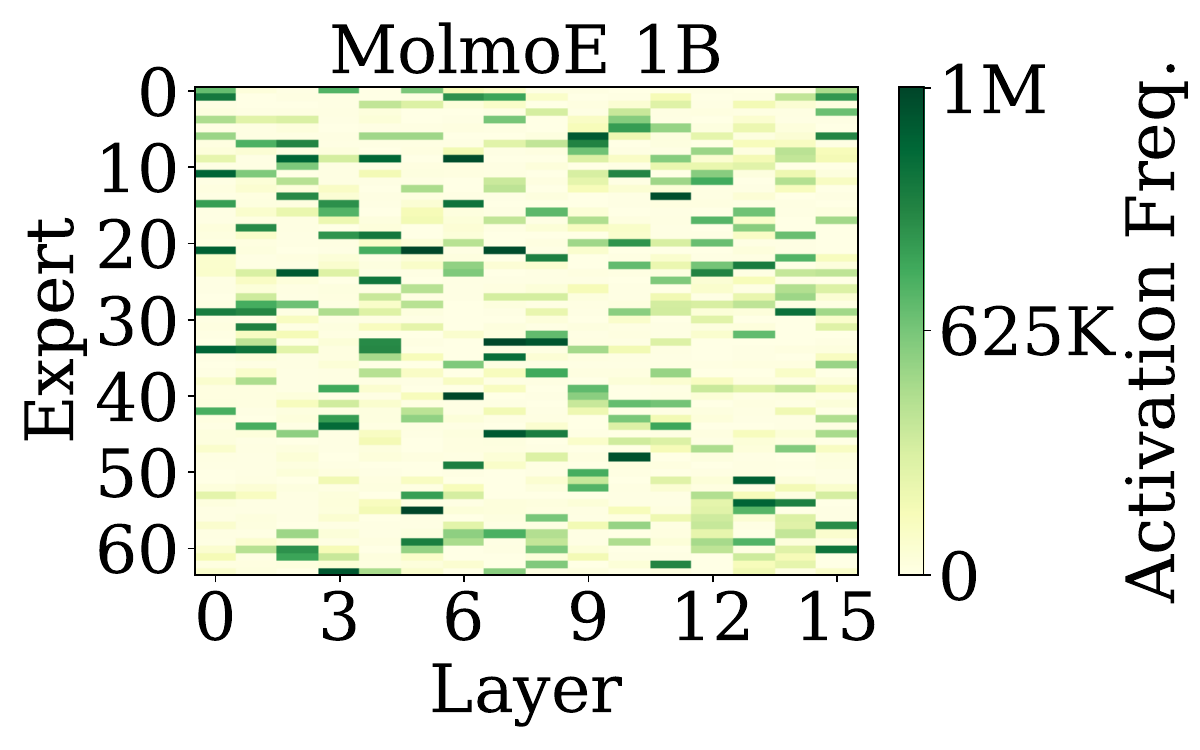}
            \label{subfig:molmoe_per_layer_freq}
        }
        \subfloat[DeepSeek VL2-Tiny]{
            \includegraphics[width=.23\linewidth]{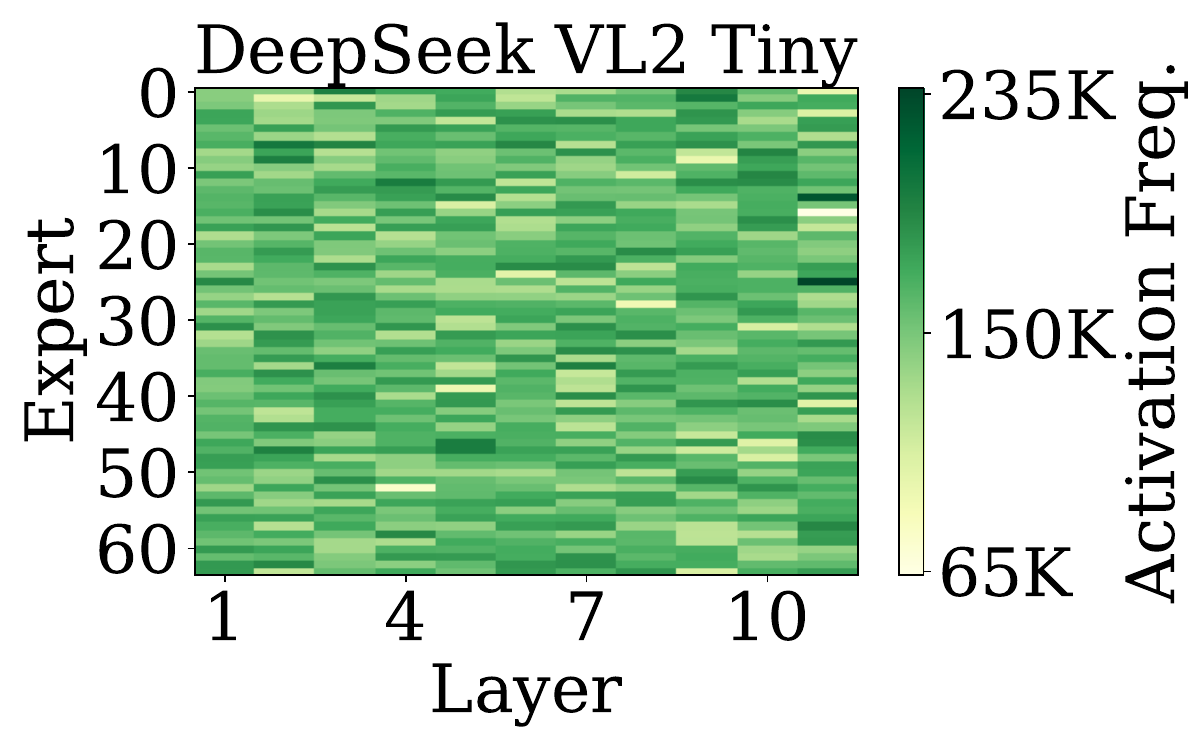}
            \label{subfig:deepseek_vl2_tiny_per_layer_freq}
        }
        \subfloat[DeepSeek VL2-Small]{
            \includegraphics[width=.23\linewidth]{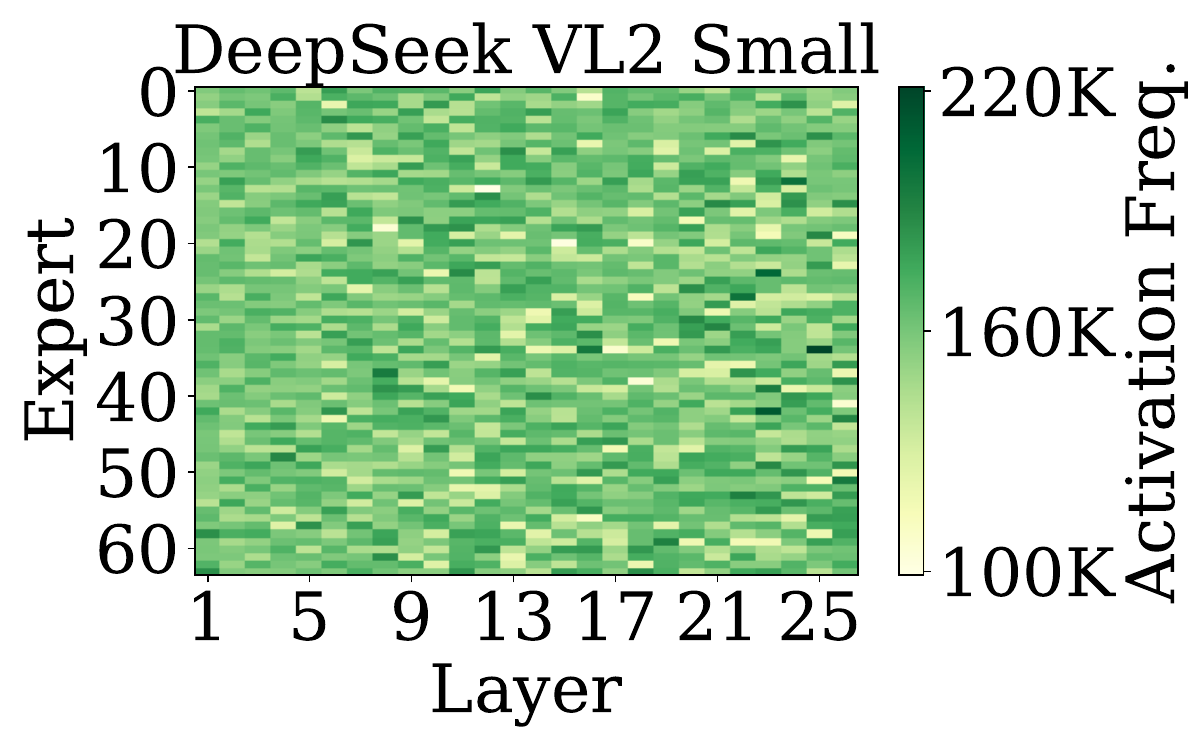}
            \label{subfig:deepseek_vl2_small_per_layer_freq}
        }
        \subfloat[DeepSeek VL2-Base]{
            \includegraphics[width=.23\linewidth]{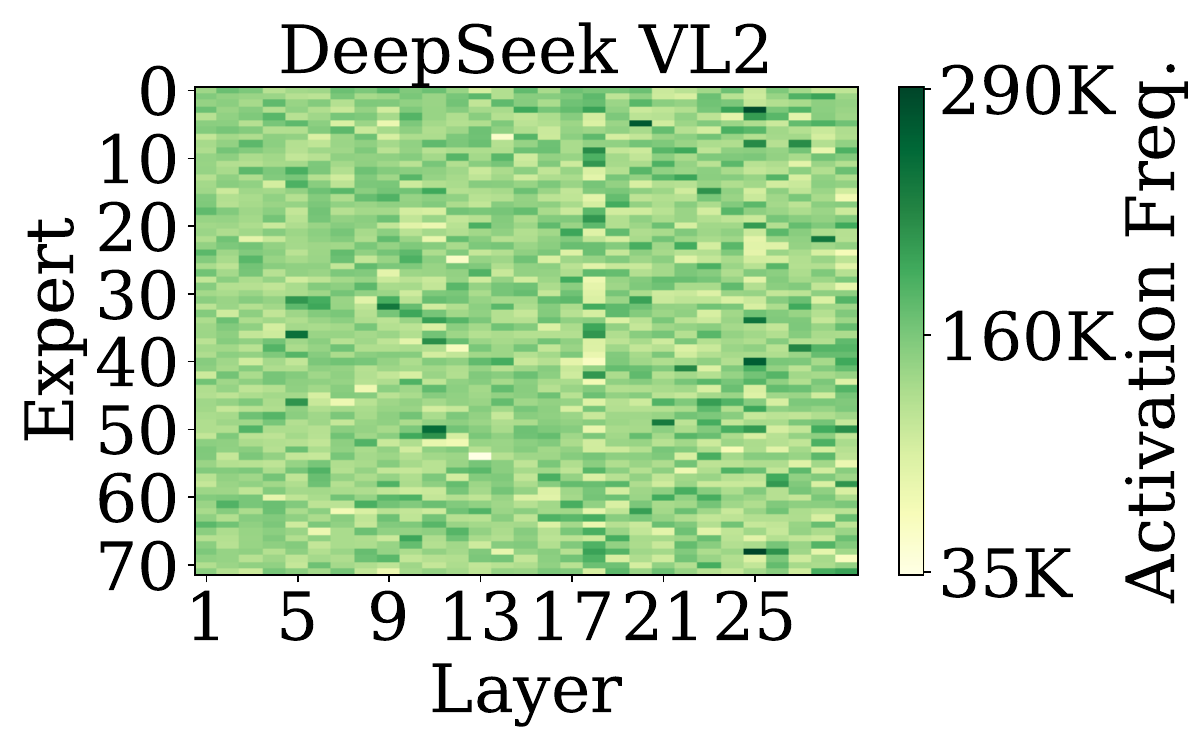}
            \label{subfig:deepseek_vl2_per_layer_freq}
        }
        \vspace{-2mm}
        \caption{Expert Activation Frequency map of MolmoE-1B and DeepSeek VL2 family Models on MME task}
        \vspace{-3mm}
        \label{fig:activation_freq}
    \end{figure*}


\section{Model Accuracy Comparison}

\subsection{Language Understanding Tasks}

\begin{figure}[H]
    \vspace{-2mm}
    \centering
    \includegraphics[width=\linewidth]{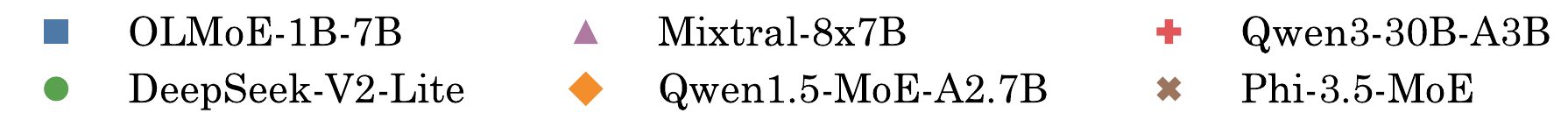}
    \begin{subfigure}{0.48\linewidth}
        \centering
        \includegraphics[width=\linewidth]{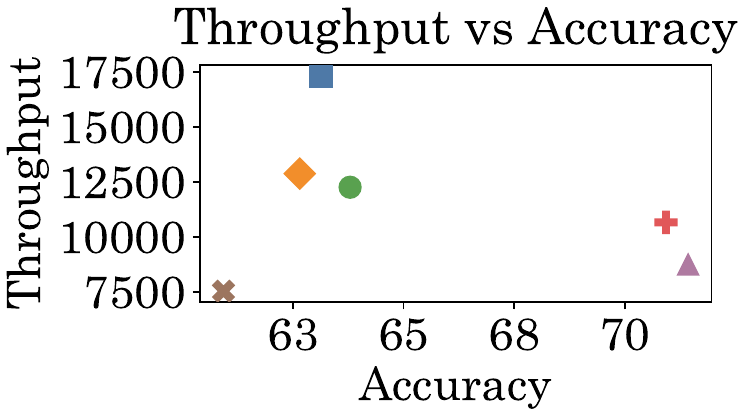}
    \end{subfigure}
    \hfill
    \begin{subfigure}{0.48\linewidth}
        \centering
        \includegraphics[width=\linewidth]{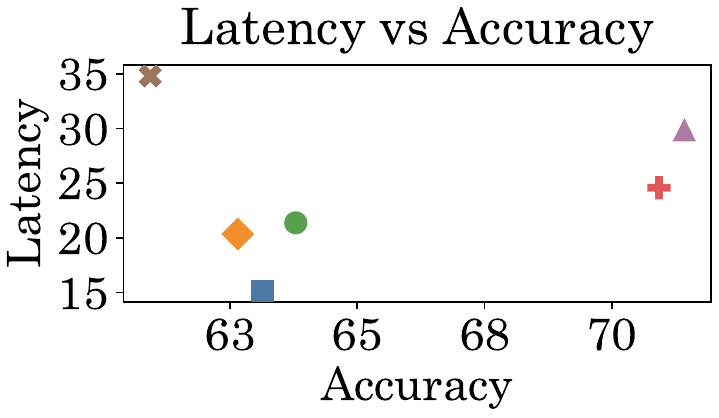}
    \end{subfigure}
    \vspace{-3mm}
    \caption{Throughput/Latency vs Accuracy for LLMs}
    \label{fig:LLM_Evaluation}
    \vspace{-5mm}
\end{figure}

We benchmark LLMs on nine widely adopted language understanding tasks from the \texttt{lm-eval} \cite{eval-harness} suite: \textit{ARC-c} \cite{Clark2018ThinkYH}, \textit{ARC-e} \cite{Clark2018ThinkYH}, \textit{BoolQ} \cite{clark2019boolq}, \textit{HellaSwag} \cite{zellers2019hellaswag}, \textit{MMLU} \cite{hendryckstest2021}, \textit{OpenBookQA} \cite{OpenBookQA2018}, \textit{RTE} \cite{wang-etal-2018-glue}, \textit{WinoGrande} \cite{sakaguchi2019winogrande}. Figure \ref{fig:LLM_Evaluation} compares throughput, latency, and average accuracy (across the all the lm-eval tasks) across six LLMs, revealing distinct trade-offs between efficiency and performance. OLMoE-1B-7B achieves the highest throughput, over 40\% higher than the next best model, while maintaining lower accuracy than MoE models such as Mixtral-8x7B and Qwen3-30B-A3B. Conversely, Qwen3-30B-A3B and Mixtral-8x7B deliver the highest accuracies but incur 60–100\% higher latency and 30–50\% lower throughput than the most efficient models. Medium-sized MoE variants like DeepSeek-V2-Lite and Qwen1.5-MoE-A2.7B lie in a balanced region, with moderate accuracy and efficiency. Phi-3.5-MoE exhibits the lowest throughput and highest latency despite competitive accuracy. These results highlight a clear performance–efficiency frontier, where small models excel in throughput and latency, while large MoEs dominate accuracy at the cost of runtime efficiency.

\subsection{Vision Language Model Tasks}

\begin{figure}[H]
    \vspace{-2.5mm} 
    \centering
    \includegraphics[width=\linewidth]{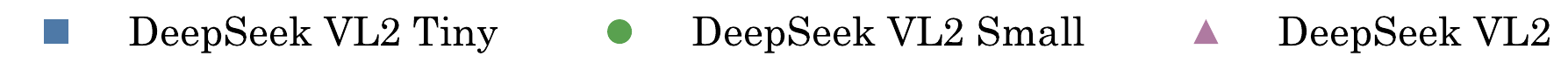}
    \begin{subfigure}{0.48\linewidth}
        \centering
        \includegraphics[width=\linewidth]{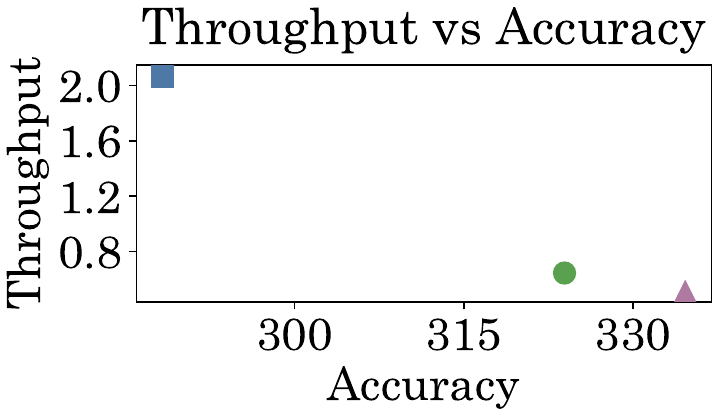}
    \end{subfigure}
    \hfill
    \begin{subfigure}{0.48\linewidth}
        \centering
        \includegraphics[width=\linewidth]{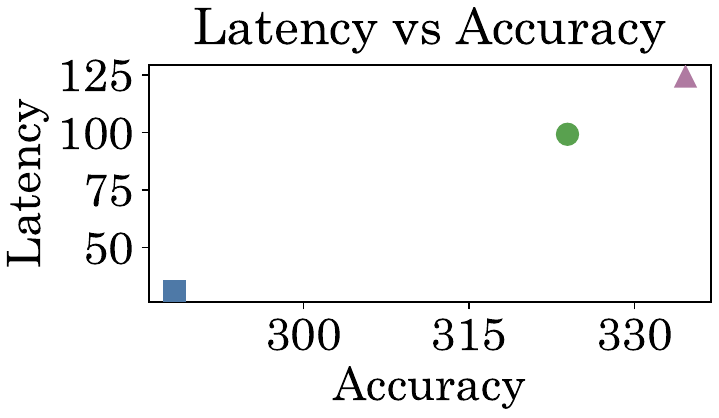}
    \end{subfigure}
    \vspace{-2mm}
    \caption{Throughput/Latency vs Accuracy for VLMs}
    \label{fig:VLM_Evaluation}
    \vspace{-4mm}
\end{figure}

We evaluate VLMs on datasets and tasks from VLMEvalKit \cite{duan2024vlmevalkit}: MME \cite{yin2023survey}, 
TextVQA \cite{singh2019towards}, AI2D \cite{kembhavi2016diagram}, DocVQA \cite{mathew2021docvqa}, MMMU \cite{yue2024mmmu}, InfoVQA \cite{mathew2022infographicvqa}, RealWorldQA \cite{zhang2024mme}, ScienceQA \cite{lu2022learn}. Figure \ref{fig:VLM_Evaluation} compares throughput and latency against average accuracy for all the tasks for the DeepSeek-VL2 Tiny, Small, and Base models. DeepSeek-VL2-Tiny achieves the highest throughput but the lowest accuracy, highlighting its suitability for speed-critical applications with reduced precision requirements. Conversely, DeepSeek VL2 delivers the highest accuracy but suffers from the lowest throughput and highest latency, making it more appropriate for accuracy-focused scenarios. DeepSeek VL2 Small offers a balanced trade-off, with moderate accuracy, throughput, and latency, serving as a middle ground between the Tiny and Base variants. This trend underscores the inherent trade-off between computational efficiency and predictive performance in VLMs.

\subsection{Expert Activation Frequency Study}

Figure \ref{fig:activation_freq} depicts the expert activation frequency (number of times each expert is selected during inference) heatmap for the DeepSeek-VL2 family and MolmoE-1B models on the MME task dataset \cite{fu2023mme}. DeepSeek-VL2 family models show a relatively uniform activation pattern across experts and layers, whereas MolmoE-1B exhibits a more sparse activation pattern, with certain experts being triggered far more often. The activation frequency in MolmoE-1B reaches up to 1M for specific experts, in contrast to DeepSeek-VL2 models, which peak around 290K. This difference arises because DeepSeek-V2 \cite{liu2024deepseek} incorporates an auxiliary loss during training to balance expert utilization, ensuring that all experts are activated more evenly. Consequently, activation frequency alone is not a dependable metric for assessing expert importance in well-balanced models. 


\section{Conclusion}
This paper introduces MoE-Inference-Bench, a comprehensive benchmarking suite that systematically evaluates the inference performance of several state-of-the-art Mixture of Experts (MoE) models across diverse hardware and algorithm optimization strategies. Through extensive evaluation of MoE models ranging from 2B to 70B parameters, mainly on Nvidia H100 GPU, we demonstrate that hyperparameter configuration, and algorithmic optimizations significantly impact MoE inference efficiency. Key findings reveal that the Nvidia H100 delivers superior performance with FP8 quantization, providing 20-30\% throughput improvements over FP16, active expert count represents the primary optimization lever with single-expert configurations achieving 50-80\% higher throughput, and vision-language models exhibit substantially larger latencies compared to text-only models.

\newpage
\section*{Acknowledgements}

This research used resources of the Argonne Leadership
Computing Facility, a U.S. Department of Energy (DOE)
Office of Science user facility at Argonne National Laboratory
and is based on research supported by the U.S. DOE Office
of Science-Advanced Scientific Computing Research Program,
under Contract No. DE-AC02-06CH11357


\bibliographystyle{ACM-Reference-Format}
\bibliography{main}





\end{document}